\documentclass[iicol, sn-basic]{sn-jnl}
\usepackage{graphicx}%
\usepackage{multirow}%
\usepackage{amsmath,amssymb,amsfonts}%
\usepackage{amsthm}%
\usepackage{mathrsfs}%
\usepackage{xcolor}%
\usepackage{textcomp}%
\usepackage{manyfoot}%
\usepackage{booktabs}%
\usepackage{pifont}
\newcommand{\cmark}{\ding{51}} 
\newcommand{\xmark}{\ding{55}} 
\usepackage{algorithm}%
\usepackage{algorithmicx}%
\usepackage{algpseudocode}%
\usepackage{listings}%
\usepackage{diagbox}
\usepackage{algorithm}
\usepackage{algpseudocode}
\usepackage{subfigure}
\usepackage{multicol}
\usepackage{makecell}
\usepackage{colortbl} 
\usepackage{arydshln} 
\usepackage{wrapfig}
\usepackage{amsmath}
\usepackage{amsthm}
\usepackage{graphicx}
\usepackage{xcolor}
\usepackage{booktabs}
\usepackage{pifont}

\definecolor{mygray}{gray}{0.92}

\begin{document}

\title[IMAGHarmony: Controllable Image Editing with Consistent Object Quantity and Layout]{IMAGHarmony: Controllable Image Editing with Consistent Object Quantity and Layout}

\author[1]{\fnm{Fei} \sur{Shen}}\email{ shenfei29@nus.edu.sg}

\author[2]{\fnm{Yutong} \sur{Gao}}\email{923106840218@njust.edu.cn}

\author[2]{\fnm{Jian} \sur{Yu}}\email{jianyu@njust.edu.cn}

\author[2]{\fnm{Xiaoyu} \sur{Du}}\email{duxy@njust.edu.cn}

\author*[3]{\fnm{Jinhui} \sur{Tang}}\email{jinhuitang@njust.edu.cn}

\affil[1]{\orgdiv{NExT++ Research Centre}, \orgname{National University of Singapore}, \country{Singapore}}

\affil[2]{ \orgname{Nanjing University of Science and Technology}, \orgaddress{\city{Nanjing}, \country{China}}}

\affil[3]{\orgname{Nanjing Forestry University}, \orgaddress{\city{Nanjing}, \country{China}}}

\vspace{-0.5cm}

\abstract{
Despite advances in diffusion-based image editing, manipulating multi-object scenes remains challenging. Existing approaches often achieve semantic changes at the expense of structural consistency, failing to preserve exact object counts and spatial layouts without introducing unintended relocations or background modifications. 
To address this limitation, we introduce quantity-and-layout-consistent image editing (QL-Edit) to modify object semantics while maintaining the original instance cardinality and spatial layout.
We propose IMAGHarmony, a parameter-efficient framework featuring a harmony-aware (HA) module that incorporates perception cues from the reference image into the diffusion process. This enables the model to jointly reason about object semantics, counts, and spatial positions for improved structural consistency.
Furthermore, we introduce a preference-guided noise selection (PNS) strategy that identifies favorable initialization conditions, substantially improving generation stability in challenging multi-object scenarios.
To support systematic evaluation, we construct HarmonyBench, a benchmark designed to measure semantic editing accuracy and structural consistency under quantity and layout constraints. 
Extensive experiments demonstrate that IMAGHarmony consistently outperforms existing methods in both structural preservation and semantic accuracy. Notably, our framework is highly efficient, requiring only 200 training images and 10.6M trainable parameters. Code, models, and data are available at \url{https://github.com/muzishen/IMAGHarmony}.
}


\maketitle

\section{Introduction}
Recent advances in diffusion models~\cite{dhariwal2021diffusion, ramesh2022hierarchical, rombach2022high, ye2023ip} have substantially improved image editing, enabling high-fidelity content manipulation under textual guidance~\cite{mou2024diffeditor, nguyen2024edit}. Existing approaches generally fall into two paradigms: mask-free~\cite{zhang2023magicbrush,brooks2023instructpix2pix,tsaban2023ledits} methods that apply global instruction-driven edits, and mask-based~\cite{brack2024ledits++,zhang2025context,yang2023paint} methods that restrict modifications to user-specified regions. Although these approaches have shown strong performance in relatively simple editing settings, reliable editing in multi-object scenes remains challenging when the edited result is required to preserve both object quantity and spatial layout from a reference image. Such constraints are important in applications where semantic editing should not alter the original visual composition, such as product editing, educational content creation, and structured scene manipulation.

\begin{figure}[t]
    \centering
    \includegraphics[width=1.0\linewidth]{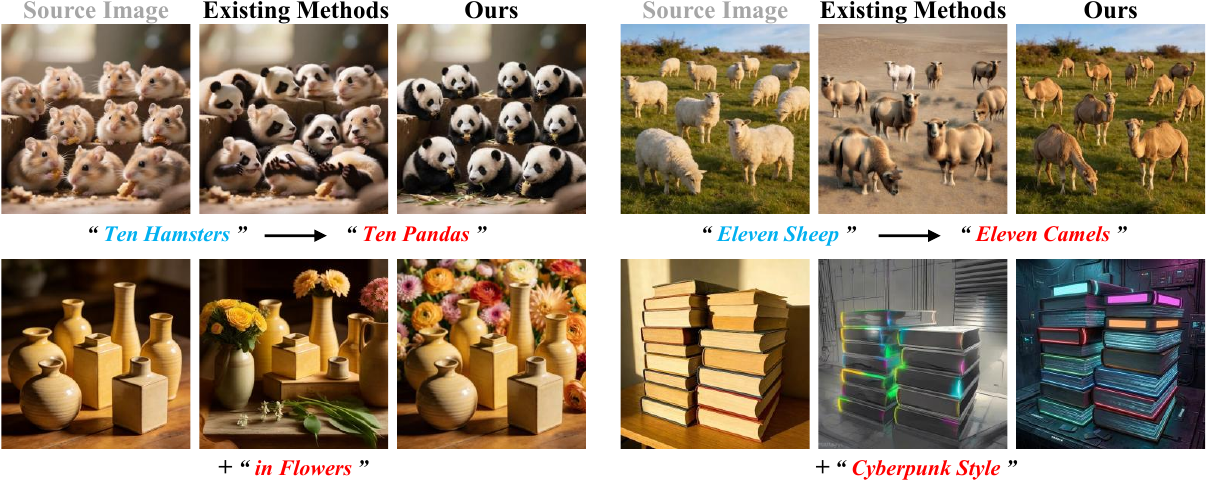}
    \vspace{-0.3cm}
    \caption{\textbf{Quantity-and-Layout-Consistent Image Editing (QL-Edit).} Unlike existing methods that suffer from cardinality drift and layout collapse (e.g., missing objects or semantic blending), IMAGHarmony precisely preserves instance counts and spatial configurations across diverse editing scenarios, including class (top), scene (bottom-left), and style (bottom-right) modifications.}
    \label{fig1}
    \vspace{-0.5cm}
\end{figure}

In such scenarios, existing methods~\cite{liu2025step1x,yu2025anyedit,feng2025dit4edit,avrahami2025stable,jiang2025vace} often achieve semantic changes at the cost of structural inconsistency. As illustrated in Fig.~\ref{fig1}, they may produce incorrect instance counts (e.g., missing animals), object relocation, semantic blending across multiple entities, or unintended modifications to surrounding regions and backgrounds. 
These failures~\cite{couairon2022diffedit,miyake2025negative,wang2026point2pix} become more pronounced when several objects of similar appearance coexist in the same image, since the model must simultaneously determine what to change, how many instances should remain, and where the edited objects should be placed. 
As a result, preserving instance cardinality and spatial arrangement during editing remains insufficiently addressed by current image editing frameworks.

Besides this structural difficulty, diffusion-based editing is also sensitive to the choice of initialization noise during inference~\cite{guo2024initno,li2024enhancing,mao2023guided}. Since the denoising trajectory depends on the starting seed~\cite{song2023consistency,yang2023diffusion}, different initializations can lead to noticeably different semantic outcomes and spatial layouts even under the same instruction. While prior studies have explored noise selection mainly from the perspective of generation quality~\cite{guo2024initno,xu2024assessing}, we find that initialization is particularly important for structure-preserving editing in multi-object scenes, where unsuitable seeds can amplify count errors, overlap among instances, or layout disruption.

To address these challenges, we formulate quantity-and-layout-consistent image editing, termed QL-Edit, a structured editing setting in which semantic modification should be achieved while maintaining the instance cardinality and coarse spatial arrangement inherited from the source image. Building on this formulation, we propose IMAGHarmony, a parameter-efficient and mask-free framework for QL-Edit. At its core is a harmony-aware (HA) module that incorporates reference-image perception cues into the diffusion process, enabling the model to jointly reason about object semantics, counts, and positions. This design improves structural consistency while preserving the editing flexibility of the underlying diffusion model.

While the HA module provides robust structural conditioning, the highly stochastic nature of diffusion sampling can still lead to topological failures if the initial noise severely conflicts with the target layout. To bridge this gap, we further introduce a preference-guided noise selection (PNS) strategy to improve generation stability in challenging multi-object editing scenarios. Rather than treating all initializations as equally suitable, PNS identifies more favorable initialization conditions for structure-preserving denoising, thereby reducing the risk of missing objects, semantic entanglement, and layout collapse. To support systematic evaluation, we construct HarmonyBench, a benchmark covering diverse quantity and layout control scenarios for multi-object editing.

Extensive experiments show that IMAGHarmony consistently outperforms existing methods in both structural preservation and semantic editing accuracy. Notably, our framework remains highly efficient, requiring only 200 training images and 10.6M trainable parameters.

Our main contributions are summarized as follows:
\begin{itemize}
    \setlength{\itemsep}{0pt}
    \setlength{\parsep}{0pt}
    \setlength{\parskip}{0pt}
    \item We formulate quantity-and-layout-consistent image editing (QL-Edit), a structured multi-object editing setting that emphasizes semantic modification while preserving instance cardinality and coarse spatial arrangement from the source image.
    \item We propose IMAGHarmony, a parameter-efficient and mask-free framework with a harmony-aware (HA) module that injects reference-image perception cues into the denoising process to improve object-level semantic and structural consistency.
    \item We devise a preference-guided noise selection (PNS) strategy that identifies more favorable initialization conditions for structure-preserving editing, improving stability in challenging multi-object scenarios.
    \item We construct HarmonyBench and demonstrate that IMAGHarmony achieves strong performance in both structural consistency and semantic accuracy, while maintaining remarkable efficiency (requiring only 200 training images and 10.6M trainable parameters).
\end{itemize}

\begin{figure*}[t]
  \centering
  \includegraphics[width=0.95\linewidth]{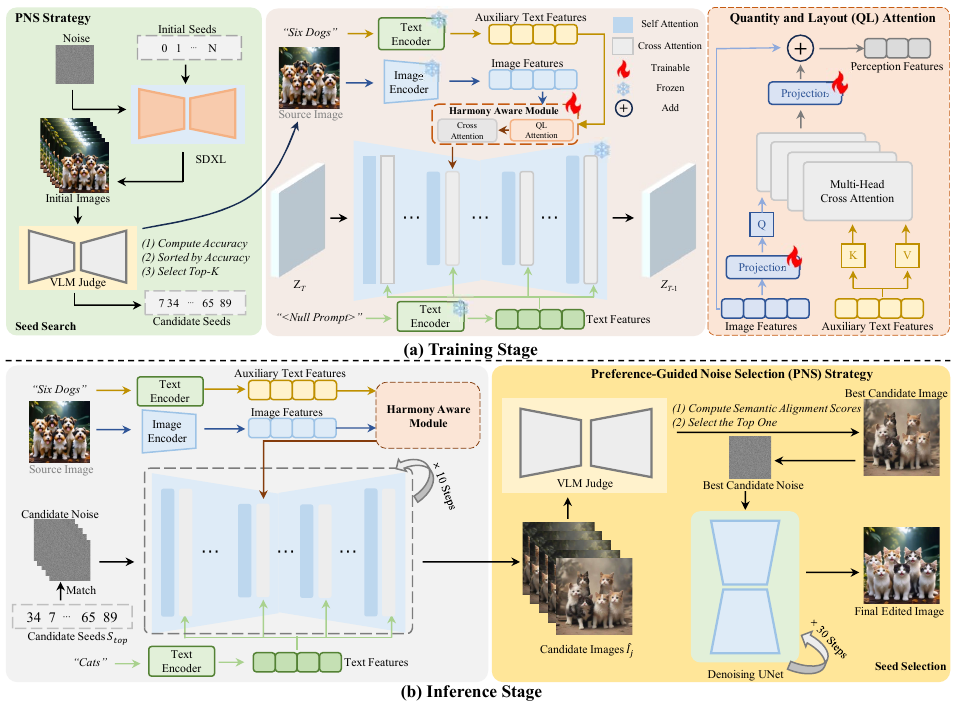}
  \caption{\textbf{Overview of the IMAGHarmony framework.} Our framework employs a \textbf{Preference-guided Noise Selection (PNS)} strategy to sample and identify the most semantically aligned noise seeds via a Vision-Language Model (VLM). The \textbf{Harmony-Aware (HA)} module then fuses multi-modal features to jointly model instance cardinality and spatial arrangements. These structural priors are injected into the frozen diffusion backbone through a cross-attention mechanism, ensuring quantity-and-layout-consistent editing without modifying the base model's weights.}
  \label{fig:framework}
  \vspace{-0.3cm}
\end{figure*}

\section{Related Work}\label{sec:rw}

\subsection{Controllable Image Editing with Diffusion Models}
Diffusion models~\cite{dhariwal2021diffusion,rombach2022high,ye2023ip} have substantially advanced text-guided image editing, with existing approaches broadly falling into mask-free~\cite{zhang2023magicbrush,brooks2023instructpix2pix,hertz2022prompt,huang2024smartedit} and mask-based~\cite{brack2024ledits++,zou2024towards,regev2025click2mask,yang2023paint} paradigms. Mask-free methods typically perform global instruction-driven edits without requiring user-specified masks, but they often provide limited control over object-level structure. Mask-based methods improve localization by restricting edits to selected regions, yet they still rely heavily on the generative model to preserve semantic and structural consistency within and around the edited area.

While these approaches~\cite{mokady2023null,kuprashevich2026nohumansrequired} have shown strong performance in relatively simple editing settings, maintaining structural consistency in multi-object scenes remains challenging. Recent studies have shown that diffusion models still exhibit limited capability in object counting and instance-aware generation~\cite{zeng2025yolo,binyamin2024make,kang2025counting}, which often leads to count drift, semantic entanglement, or object relocation when multiple entities must be edited simultaneously. As a result, existing editing frameworks~\cite{brooks2023instructpix2pix,zhang2023magicbrush,zou2024towards} may struggle to preserve instance cardinality and spatial arrangement in reference-guided multi-object editing. In contrast, our work focuses on quantity-and-layout-consistent image editing (QL-Edit), where semantic modification is achieved while maintaining the object quantity and coarse spatial structure inherited from the source image, notably accomplishing this in a fully mask-free manner.

\subsection{Semantic Guidance and Preference Modeling with Vision-Language Models}
Vision-language models (VLMs)~\cite{chiang2023vicuna} have become increasingly important for improving semantic controllability in image generation and editing~\cite{han2024proxedit,mou2023dragondiffusion}. Early studies~\cite{qu2023layoutllm,feng2023layoutgpt} explored the use of large language models for explicit layout planning, while more recent unified generative frameworks~\cite{wang2024emu3,ge2024seed,xie2024show} attempt to model text and image generation within a shared multimodal space. VLMs have also been used as scoring or preference models~\cite{xu2024assessing,kirstain2023pick} to evaluate generation quality, rerank candidate outputs, or select more favorable initializations~\cite{guo2024initno}.

Despite their promise, existing VLM-guided pipelines~\cite{bai2025uniedit,kumari2025learning,wang2025seededit,li2025edit,jadhav2025towards} often require explicit structural annotations such as bounding boxes, depend on computationally intensive unified architectures, or mainly apply preference modeling exclusively at inference time. Different from these directions, IMAGHarmony is highly parameter-efficient and does not rely on explicit layout supervision. Instead, it injects reference-image perception cues into the diffusion process to improve object-level semantic and structural consistency. Furthermore, unlike prior works that rely on post-hoc reranking, our preference-guided noise selection (PNS) strategy operates across both data preparation and inference stages, fundamentally stabilizing the structure-preserving generation trajectories.

\section{Method}\label{sec:method}
In Section~\ref{subsec:problem_difinition}, we first formulate quantity-and-layout-consistent image editing (QL-Edit), clarify its distinction from related image editing settings, and introduce the notation used throughout the paper. We then present IMAGHarmony, a framework tailored to QL-Edit. Section~\ref{subsec:overall} and Fig.~\ref{fig:framework} provide an overview of the full pipeline. Specifically, IMAGHarmony augments a frozen Stable Diffusion XL (SDXL) denoising UNet with a harmony-aware (HA) module, which is detailed in Section~\ref{subsec:HA}. Section~\ref{subsec:PNS} further introduces a preference-guided noise selection (PNS) strategy that improves stability in challenging multi-object editing scenarios. Finally, Section~\ref{subsec:train} summarizes the complete training and inference procedures.


\subsection{Preliminaries}\label{subsec:problem_difinition}

\noindent\textbf{Problem Description.}
We formulate quantity-and-layout-consistent image editing, termed QL-Edit, as a structured editing setting in which textual instructions are applied while maintaining the object quantity and coarse spatial arrangement of a reference image. Unlike conventional image editing tasks that mainly emphasize semantic modification or stylistic transformation, QL-Edit places explicit emphasis on structural consistency, requiring the edited result to preserve instance cardinality and the overall spatial organization inherited from the source image.
We consider three representative forms of QL-Edit:
(1) \underline{Class Editing}: modifying the object category while preserving the original quantity and layout (e.g., ``Six Tigers $\rightarrow$ Six Lions'', where the semantic category changes but the number and arrangement of instances remain consistent; see Fig.~\ref{fig1}, second row);
(2) \underline{Scene Editing}: modifying the background or contextual environment while maintaining the object quantity and spatial arrangement (e.g., adding ``flowers in the background''; see Fig.~\ref{fig1}, third row, left);
(3) \underline{Style Editing}: applying a global stylistic transformation without changing the semantic content, object quantity, or spatial configuration (e.g., adding ``cyberpunk style''; see Fig.~\ref{fig1}, third row, right).
Therefore, QL-Edit focuses on reference-guided multi-object editing under quantity and layout constraints, rather than unconstrained instruction-based image manipulation.


\noindent\textbf{Symbol Definition.}
For clarity, we summarize the main symbols used in IMAGHarmony in Table~\ref{tab:symbols}.

\begin{table}[t]
    \renewcommand{\arraystretch}{0.9}
    \setlength{\tabcolsep}{4pt}
    \centering
    \caption{Definitions of main symbols used in this paper.}
    \begin{tabular}{c|p{5.8cm}}
        \hline
        Notation & Definition \\ \hline
        $\boldsymbol{z}_0$ & Ground-truth latent representation \\
        $\boldsymbol{\epsilon}$ & Gaussian noise \\
        $t$ & Timestep \\
        $T$ & Total number of denoising steps \\
        $\boldsymbol{\epsilon}_\theta$ & Diffusion model noise predictor \\
        $w$ & Guidance scale for inference \\
        $\alpha$ & Scaling factor for QL attention \\
        $\beta$ & Scaling factor for perception cross-attention \\
        $\boldsymbol{z}_t$ & Noisy latent at step $t$ \\
        $\boldsymbol{F}_v$ & Visual feature from reference image \\
        $\boldsymbol{F}_a$ & Auxiliary textual feature \\
        $\boldsymbol{F}_t$ & Textual feature from editing instruction \\
        $\boldsymbol{F}_p'$ & Projected perception feature \\
        $\mathcal{S}_{\text{top}}$ & Candidate seed set \\
        $\boldsymbol{\hat{I}}_j$ & Candidate image \\
        $A_i$ & Binary accuracy score \\
        $R_j$ & Semantic alignment score \\
        \hline
    \end{tabular}
    \label{tab:symbols}
    \vspace{-0.5cm}
\end{table}

\subsection{Overall Framework}\label{subsec:overall}
As illustrated in Fig.~\ref{fig:framework}, IMAGHarmony consists of two key components: a harmony-aware (HA) module for quantity- and layout-consistent conditioning, and a preference-guided noise selection (PNS) strategy for improving initialization stability.

Given an editing instruction and a reference image, we first encode the auxiliary text and the source image using frozen text and image encoders, respectively, to obtain textual and visual features. These features are then processed by the HA module to produce harmony features that explicitly encode object quantity and implicitly capture spatial structure. The resulting features are injected into a frozen Stable Diffusion XL (SDXL) denoising UNet through an additional learnable cross-attention branch, while the original frozen cross-attention branch is retained to preserve the generative prior of the backbone model. Following prior visual conditioning designs, the new cross-attention branch is inserted only into the ``Down4'' block of the UNet.

To further improve stability in challenging multi-object editing scenarios, we introduce PNS, which identifies more favorable initialization conditions for structure-preserving denoising. During data preparation, we pre-screen a small set of candidate seeds that are more likely to satisfy quantity-related constraints. During inference, we perform shallow denoising on these candidates under the editing instruction, evaluate their consistency, and then select the most suitable initialization for full denoising. This design reduces the risk of count errors, semantic entanglement, and layout disruption. The complete training and inference procedures are summarized in Section~\ref{subsec:train}.

\subsection{Harmony Aware Module}
\label{subsec:HA}

\noindent\textbf{Motivation.}
Existing studies~\cite{zeng2025yolo,binyamin2025make} suggest that diffusion models remain limited in instance-aware structural control, especially when multiple objects must be edited while preserving quantity and layout. In such cases, visual features alone may be insufficient to reliably encode both instance cardinality and spatial organization. To address this issue, we propose a harmony-aware (HA) module that combines auxiliary textual cues and reference-image visual cues to improve quantity and layout consistency during editing.

\noindent\textbf{Architecture.}
As shown in Fig.~\ref{fig:framework}(a), the HA module consists of a quantity-and-layout (QL) attention block followed by a cross-attention injection mechanism. The QL attention block contains two learnable linear projections and a multi-head cross-attention module, which enhances cross-modal interaction between structural textual cues and visual perception features.

Let $F_v \in \mathbb{R}^{B \times N_v \times D_v}$ and $F_a \in \mathbb{R}^{B \times N_t \times D_t}$ denote the visual features and auxiliary textual features, respectively, where the auxiliary text is derived from an explicit count-object prompt such as ``Six dogs''. Here, $B$ is the batch size, $N_v$ and $N_t$ are the sequence lengths of visual and textual tokens, and $D_v$ and $D_t$ are the corresponding feature dimensions. We first project the visual feature $F_v$ into the auxiliary textual feature space:
\begin{equation}
F_v' = F_v W_{\text{proj}},
\label{eq:proj}
\vspace{-0.3cm}
\end{equation}
where $F_v' \in \mathbb{R}^{B \times N_v \times D_t}$ and $W_{\text{proj}} \in \mathbb{R}^{D_v \times D_t}$ is a learnable projection matrix.

Next, we use $F_v'$ as the query and $F_a$ as both key and value in a multi-head cross-attention module to capture quantity- and layout-related dependencies:
\begin{equation}
F_{va} = \text{MultiHead}(F_v', F_a, F_a).
\label{eq:mh_input}
\vspace{-0.3cm}
\end{equation}
We then project the fused features back into the original visual feature space:
\begin{equation}
F_{va}' = F_{va} W_{\text{align}},
\label{eq:proj2}
\vspace{-0.3cm}
\end{equation}
where $F_{va}' \in \mathbb{R}^{B \times N_v \times D_v}$ and $W_{\text{align}} \in \mathbb{R}^{D_t \times D_v}$ is a learnable projection matrix.

To control the contribution of the injected structural cues while preserving the original visual representation, we introduce a scaling hyperparameter $\alpha$ and define the final perception feature $F_p \in \mathbb{R}^{B \times N_v \times D_v}$ as
\begin{equation}
F_p = \alpha F_{va}' + F_v.
\label{eq:hf}
\vspace{-0.3cm}
\end{equation}
The resulting feature $F_p$ provides a harmonized representation that integrates object quantity and spatial structure with the reference-image visual features.

To inject $F_p$ while preserving the original SDXL architecture, we adopt a dual-branch cross-attention design. One branch keeps the frozen cross-attention module from SDXL, conditioned on \texttt{<Null Prompt>} to avoid redundant semantic interference. The other branch uses the perception feature to introduce quantity- and layout-aware conditioning. Following the design in Fig.~\ref{fig:framework}(a), this additional branch is inserted only into the ``Down4'' block of the UNet.

To remain compatible with the pretrained visual conditioning space, the perception feature $F_p$ is first mapped through a frozen projection layer inherited from IP-Adapter, yielding $F_p' = \text{Proj}_{\text{IP}}(F_p)$. This design avoids introducing extra trainable parameters in the modality alignment stage and leverages the pretrained alignment between visual features and diffusion conditioning. The projected feature $F_p'$ is then fed into a learnable cross-attention layer to interact with the query features. Let $F_t \in \mathbb{R}^{B \times N_t \times D_t}$ denote the textual feature. The final attention output is computed as
\begin{equation}
\begin{split}
Z_{\text{new}} =& \, \text{Softmax}\left(\frac{QK_{t}^{T}}{\sqrt{d}}\right)V_{t} \\
&+ \beta \cdot \text{Softmax}\left(\frac{QK_{p}^{T}}{\sqrt{d}}\right)V_{p},
\end{split}
\label{eq:dual_attn}
\vspace{-0.3cm}
\end{equation}
where $Q = h W_q$ is the query projected from the intermediate spatial features $h$ of the UNet at the current resolution. The text branch uses frozen projections $K_t = F_t W_k^t$ and $V_t = F_t W_v^t$, while the perception branch uses $K_p = F_p' W_k^p$ and $V_p = F_p' W_v^p$ with newly introduced trainable weights. The hyperparameter $\beta$ balances the contribution of the perception branch. This dual-branch design enables IMAGHarmony to inject quantity- and layout-aware conditioning into the editing process while retaining the generalization ability of the original SDXL model.

\subsection{Preference-Guided Noise Selection Strategy}
\label{subsec:PNS}

\noindent\textbf{Motivation.}
Diffusion models are sensitive to initialization noise~\cite{guo2024initno,xu2024assessing}, and this sensitivity becomes more pronounced in multi-object editing settings where quantity and layout consistency must be preserved. Even under the same editing instruction, different seeds may lead to noticeably different semantic outcomes and spatial arrangements. To improve stability, we introduce a preference-guided noise selection (PNS) strategy that identifies more favorable initialization conditions for structure-preserving denoising.

\noindent\textbf{Seed Search.}
As shown in Fig.~\ref{fig:framework}(a), we perform a seed search during data preparation to pre-select a small set of candidate initializations that are more likely to satisfy quantity-related constraints. Specifically, we uniformly sample $N$ random seeds $\{s_i\}_{i=1}^{N}$, and for each seed $s_i$, we generate an image $I_i$ using SDXL~\cite{podell2023sdxlimprovinglatentdiffusion} under a fixed textual prompt. To assess coarse semantic fidelity, we employ a vision-language model (VLM) with instance-specific counting queries, such as ``How many dogs are in the image?''. The predicted count is compared with the target count, yielding a binary score $A_i \in \{0,1\}$.

After computing $A_i$ for all sampled seeds, we rank the results and retain the top-$k$ seeds to form a candidate set $\mathcal{S}_{\text{top}}$. In practice, we use $N=100$ and $k=5$, which provides a good balance between efficiency and diversity. This stage serves as a coarse screening process that filters out unsuitable initializations before editing.

\noindent\textbf{Seed Selection.}
As shown in Fig.~\ref{fig:framework}(b), during inference we initialize the diffusion process with each candidate seed $s_j \in \mathcal{S}_{\text{top}}$ and perform shallow denoising for a small number of steps under the target editing instruction to obtain a candidate image $\hat{I}_j$. Each candidate image is then evaluated using CogVLM2~\cite{hong2024cogvlm2visuallanguagemodels}, which produces a semantic consistency score $R_j \in [0,1]$ measuring its alignment with the editing instruction.

We then select the seed associated with the highest-scoring candidate and use it to reinitialize the model for full denoising, yielding the final edited image. In this way, the seed search stage in data preparation performs coarse filtering based on quantity-related constraints, while the seed selection stage at inference uses a more comprehensive semantic consistency criterion under the actual editing instruction. Together, these two stages improve robustness against count errors, semantic entanglement, and layout collapse in challenging multi-object editing scenarios.

\begin{algorithm}[t]
\caption{IMAGHarmony training and inference procedure}
\label{alg:imagharmony}
\begin{algorithmic}[1]
\Require Reference image $I_s$, editing prompt $P_e$, auxiliary prompt $P_a$, candidate seeds $\mathcal{S}_{\text{top}}$
\Ensure Edited image $I_{\text{edit}}$

\State Extract $F_v$ from $I_s$ and $F_a$ from $P_a$
\State Compute $F_p$ by Eqs.~\eqref{eq:proj}--\eqref{eq:hf} \Comment{HA module}
\State Project $F_p$ to $F_p'$ via frozen IP-Adapter layer
\State Sample $t$ and $\boldsymbol{\epsilon}$, and construct noisy latent $\boldsymbol{z}_t$
\State Predict noise $\boldsymbol{\epsilon}_\theta$ with injected $F_p'$
\State Update HA by minimizing Eq.~\eqref{eq:loss}

\For{each $s_j \in \mathcal{S}_{\text{top}}$}
    \State Perform shallow denoising with Eq.~\eqref{eq:cfg}
    \State Obtain candidate image $\hat{I}_j$ and score $R_j$
\EndFor

\State Select $s^\star = \arg\max_{s_j \in \mathcal{S}_{\text{top}}} R_j$
\State Perform full denoising with $s^\star$ using Eq.~\eqref{eq:cfg}
\State \Return $I_{\text{edit}}$
\end{algorithmic}
\end{algorithm}

\subsection{Training and Inference}
\label{subsec:train}
For clarity, the overall training and inference procedures of IMAGHarmony are summarized in Algorithm~\ref{alg:imagharmony}.

\noindent\textbf{Training.}
During training, we optimize only the proposed HA module while keeping the SDXL UNet frozen, so as to preserve the pretrained generative prior of the backbone model. Given the textual feature $F_t$ and the projected perception feature $F_p'$, we adopt the standard noise-prediction objective in the latent space:
\begin{equation}
\begin{split}
L_{\text{MSE}} =& \, \mathbb{E}_{\boldsymbol{z}_0,\boldsymbol{\epsilon},F_t,F_p',t} \Big\| \boldsymbol{\epsilon} \\
&- \boldsymbol{\epsilon}_{\theta}(\boldsymbol{z}_t, F_t, F_p', t) \Big\|^2,
\end{split}
\label{eq:loss}
\end{equation}
where $\boldsymbol{z}_0$ denotes the latent representation of the ground-truth image encoded by the VAE, $\boldsymbol{\epsilon} \sim \mathcal{N}(0,I)$ is Gaussian noise, $\boldsymbol{z}_t$ is the noisy latent at timestep $t$, and $\boldsymbol{\epsilon}_{\theta}$ is the predicted noise. The timestep $t \in [0,T]$ is uniformly sampled, where $T$ denotes the total number of denoising steps. By optimizing this objective, the HA module learns to inject quantity- and layout-aware conditioning into the frozen diffusion backbone, thereby improving both semantic accuracy and structural consistency.

\noindent\textbf{Inference.}
During inference, we employ classifier-free guidance (CFG)~\cite{ho2022classifier} to strengthen conditional editing. The final noise prediction is computed as
\begin{equation}
\begin{split}
\hat{\boldsymbol{\epsilon}}_{\theta}(\boldsymbol{z}_t, F_t, F_p', t) =& \, w \cdot \boldsymbol{\epsilon}_{\theta}(\boldsymbol{z}_t, F_t, F_p', t) \\
&+ (1-w) \cdot \boldsymbol{\epsilon}_{\theta}(\boldsymbol{z}_t, \emptyset, \emptyset, t),
\end{split}
\label{eq:cfg}
\end{equation}
where $w$ is the guidance scale controlling the strength of conditional information, and $\emptyset$ denotes the null embeddings for unconditional guidance. In practice, inference is further combined with the proposed PNS strategy. Specifically, we first perform shallow denoising on a small set of candidate seeds, evaluate their consistency under the target editing instruction, and then select the most suitable initialization for full denoising. This process improves robustness in challenging multi-object editing scenarios and reduces the risk of count errors, semantic entanglement, and layout disruption.

\section{Theoretical Analysis}
\label{sec:theory}

In this section, we provide theoretical justifications for two key design choices in IMAGHarmony: injecting the Harmony-Aware (HA) module at the lowest-resolution bottleneck (Down4), and selecting diffusion seeds with the Preference-Guided Noise Selection (PNS) strategy.

\subsection{Why Injecting at Down4?}
\label{sec:theory_down4}

We first formalize the intuition that object quantity and spatial layout correspond to coarse structural attributes that are primarily encoded in low-frequency components.

Let the diffusion generator be denoted as
\begin{equation}
I = G_\theta(s; c),
\end{equation}
where $s$ is the initial noise seed and $c$ denotes the conditioning signals (including both text and perception cues). We define a structural projection operator
\begin{equation}
\Pi: \mathcal{I} \rightarrow \mathcal{S},
\end{equation}
which extracts coarse structural properties such as object counts and spatial centroids. Given a reference image $I_r$, the feasible structural set is defined as
\begin{equation}
\mathcal{C}(I_r) = \left\{ I \mid \Pi(I) \in \mathcal{S}(I_r) \right\}.
\end{equation}

It has been widely observed that diffusion U-Nets exhibit a hierarchical spectral bias~\cite{hertz2022prompt}: deeper layers (lower resolutions) primarily dictate global structures, while shallower layers synthesize high-frequency details such as textures and edges. Therefore, structural guidance injected at deeper layers explicitly influences the low-frequency layout formation.

\vspace{0.15cm}
\noindent\textbf{Proposition 1 (Structural dominance of Down4).} \textit{Under a fixed perturbation magnitude, injecting structural signals at the lowest-resolution bottleneck (Down4) maximizes the induced change in coarse structural attributes measured by $\Pi(\cdot)$.}
\vspace{0.15cm}

\noindent\textbf{Justification.}
Since deeper layers modulate low-frequency components, perturbations introduced at Down4 govern the global image layout. Because the structural projection $\Pi(\cdot)$ explicitly captures macroscopic properties (quantity and spatial arrangement), it is inherently more sensitive to low-frequency structural shifts than to high-frequency texture variations. Consequently, for an equivalent guidance budget, injecting HA at Down4 yields the most effective alignment within $\mathcal{C}(I_r)$.

\noindent\textbf{Discussion.}
This proposition conceptually validates why integrating structural cues at Down4 surpasses shallower layers for quantity- and layout-consistent editing. It also aligns with our empirical observations (e.g., Fig.~\ref{fig:diff_block}), where Down4 injection achieves the highest structural controllability while preserving image fidelity.

\subsection{Analysis of the PNS Strategy}
\label{sec:theory_pns}

We next analyze the Preference-Guided Noise Selection (PNS) strategy. The core objective is to determine whether selecting the optimal seed from a candidate pool systematically increases the probability of generating a structurally feasible result.

Let the feasibility indicator be
\begin{equation}
Y(s) = \mathbf{1}\!\left[G_\theta(s;c) \in \mathcal{C}(I_r)\right],
\end{equation}
where $Y(s)=1$ implies that the generated image satisfies the quantity and layout constraints. The baseline success probability of a single random seed is
\begin{equation}
\pi_1 = \mathbb{P}_{s}[Y(s)=1].
\end{equation}

Assume each seed $s$ is evaluated by a proxy semantic alignment score $R(s) \in [0, 1]$ (e.g., predicted by the VLM). Given a candidate pool of $K$ independent seeds $\mathcal{S}_{\text{top}} = \{s_1, \dots, s_K\}$, PNS selects the optimal seed:
\begin{equation}
s^\star = \arg\max_{s_j \in \mathcal{S}_{\text{top}}} R(s_j).
\end{equation}
Let $\pi_K$ denote the probability that the generated image using the selected seed $s^\star$ is feasible.

\vspace{0.15cm}
\noindent\textbf{Proposition 2 (Best-of-$K$ improvement).} \textit{Assuming a monotonic dependency where higher proxy scores $R(s)$ correlate with a higher feasibility probability, the success probability of the PNS strategy strictly satisfies $\pi_K \ge \pi_1, \forall K \ge 1$. Moreover, if the proxy score is an informative indicator and $K>1$, the inequality is strict.}
\vspace{0.15cm}

\noindent\textbf{Justification.}
Let $R$ be the score distribution of a randomly sampled seed, and let $M_K = \max_j R(s_j)$ be the highest score among $K$ independent candidates. According to order statistics, $M_K$ stochastically dominates $R$. By selecting the highest-scoring seed $s^\star$, we monotonically increase the expected proxy score. Under the premise that $R(s)$ is an informative proxy for structural feasibility $Y(s)$, the success probability of $s^\star$ is bounded below by that of a random seed. Thus, $\pi_K \ge \pi_1$.

\noindent\textbf{Discussion.}
This analysis demonstrates that PNS is not an arbitrary heuristic but a probabilistically sound mechanism. Through stochastic dominance, selecting the best seed from multiple candidates consistently elevates the lower bound of generation feasibility. In practice, the performance gain saturates as $K$ increases, while the inference overhead grows linearly, which justifies our empirical choice of a modest candidate pool (e.g., $K=5$).

\begin{figure}[t]
  \centering
  \includegraphics[width=0.95\linewidth]{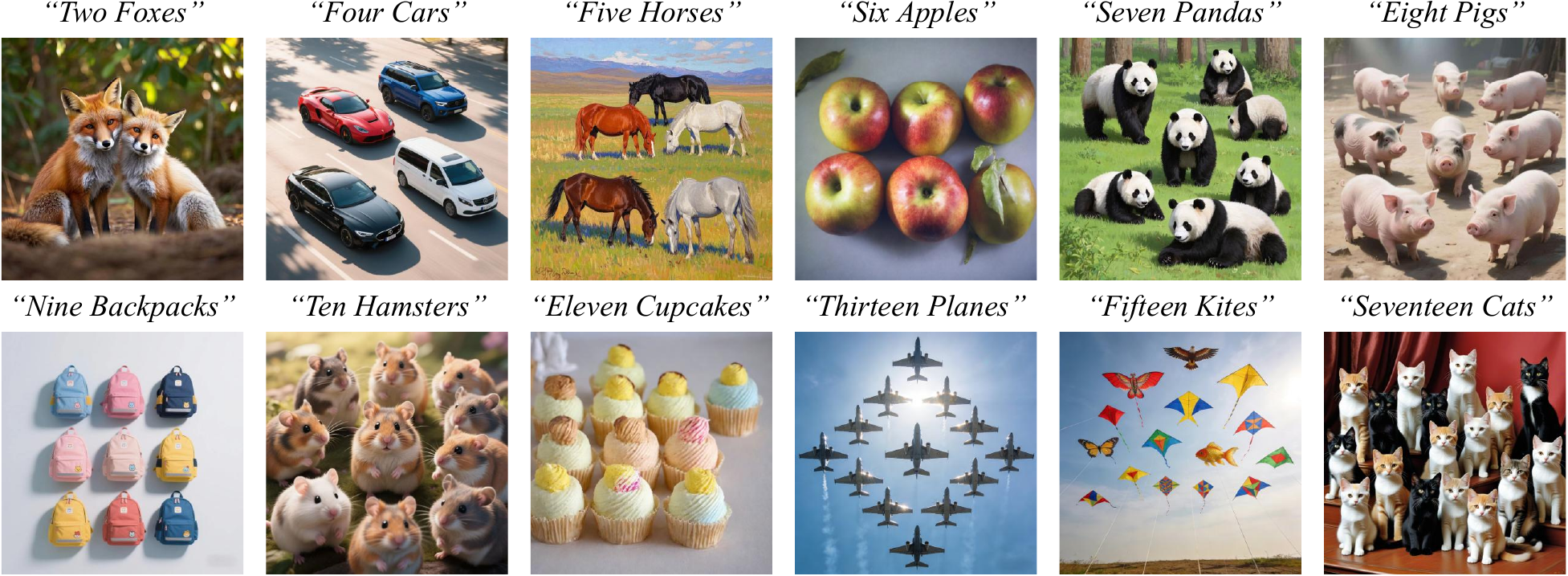}
  \caption{\textbf{Representative samples from HarmonyBench.} Each entry pairs a source image with a quantity-aware caption. To ensure rigorous ground-truth fidelity, all instance counts are cross-verified by human annotators with YOLO-World~\cite{cheng2024yoloworldrealtimeopenvocabularyobject} assistance.}
  \label{fig:example}
  \vspace{-0.3cm}
\end{figure}


\begin{table*}[t]
\centering
\setlength{\tabcolsep}{6pt}
\caption{\textbf{Quantitative comparisons on proposed HarmonyBench.} IMAGHarmony achieves the best performance.}

\label{table:comparison}
\renewcommand{\arraystretch}{1.1}
\setlength{\tabcolsep}{6pt}
\resizebox{0.9\textwidth}{!}{
\begin{tabular}{l|ccc|ccc|ccc|ccc}
\toprule
\rowcolor{mygray}
\textbf{Method} & \multicolumn{3}{c|}{\textbf{Average ($\uparrow$)}} &
\multicolumn{3}{c|}{\textbf{Class Editing ($\uparrow$)}} & 
\multicolumn{3}{c|}{\textbf{Scene Editing ($\uparrow$)}} & 
\multicolumn{3}{c}{\textbf{Style Editing ($\uparrow$)}} \\
\rowcolor{mygray}
& OCA & AP & IR & OCA & AP & IR & OCA & AP & IR & OCA & AP & IR \\
\midrule
SEED-X~\cite{ge2024seed}                   & 41.3 & 26.3 & 0.344 & 43.1 & 15.3 & 0.601 & 41.5 & 37.4 & 0.206 & 39.2 & 26.3 & 0.224 \\
Show-o~\cite{xie2024show}                  & 48.9 & 50.7 & 0.158 & 35.5 & 31.0 & 0.485 & 80.5 & 77.5 & -0.134 & 30.6 & 43.6 & 0.124 \\
Emu-3~\cite{wang2024emu3}                  & 51.6 & 44.0 & 0.271 & 51.9 & 33.9 & 0.585 & 57.7 & 66.8 & -0.041 & 45.1 & 31.2 & 0.268 \\
LEDITS~\cite{tsaban2023ledits}             & 65.6 & 59.7 & 0.025 & 45.2 & 43.1 & 0.463 & 77.2 & 71.9 & -0.135 & 74.4 & 64.1 & -0.253 \\
LEDITS++~\cite{brack2024ledits++}          & 70.0 & 66.0 & 0.046 & 48.4 & 45.4 & 0.498 & 83.5 & 88.4 & -0.187 & 77.9 & 63.9 & -0.174 \\
LoMOE~\cite{chakrabarty2024lomoe}          & 72.3 & 67.9 & 0.059 & 66.4 & 63.8 & -0.007 & 66.8 & 73.5 & 0.096 & 83.7 & 66.4 & 0.088 \\
InstructPix2Pix~\cite{brooks2023instructpix2pix} & 82.5 & 78.2 & 0.223 & 66.7 & 65.3 & 0.372 & 91.8 & 91.1 & 0.092 & 89.1 & 78.2 & 0.205 \\
MagicBrush~\cite{zhang2023magicbrush}      & 84.1 & 78.8 & 0.252 & 70.3 & 69.2 & 0.437 & 91.3 & 90.5 & 0.104 & 90.8 & 76.7 & 0.216 \\
IC-Edit~\cite{zhang2025context}            & 84.1 & 81.7 & 0.344 & 73.2 & 74.6 & 0.548 & 93.7 & 90.6 & 0.198 & 85.5 & 79.8 & 0.287 \\
Any-Edit~\cite{yu2025anyedit}              & 84.8 & 82.6 & 0.365 & 74.5 & 75.2 & 0.585 & 94.0 & 91.2 & 0.208 & 85.9 & 81.4 & 0.302 \\
\hline
\rowcolor{blue!7} {IMAGHarmony (SD1.5)}    & {88.3} & {85.5} & {0.401} & {79.6} & {75.6} & {0.686} & {94.1} & {91.4} & {0.204} & {91.2} & {89.4} & {0.312} \\
\rowcolor{blue!7} \textbf{IMAGHarmony (SDXL)} & \textbf{88.8} & \textbf{85.9} & \textbf{0.408} & \textbf{80.7} & \textbf{75.7} & \textbf{0.693} & \textbf{94.3} & \textbf{91.8} & \textbf{0.217} & \textbf{91.4} & \textbf{90.2} & \textbf{0.315} \\
\bottomrule
\end{tabular}}
\end{table*}

\begin{table*}[t]
\centering
\setlength{\tabcolsep}{6pt}
\caption{\textbf{Quantitative comparisons on proposed HarmonyBench} according to the number of the object.}

\label{table:comparison_num}
\renewcommand{\arraystretch}{1.1}
\setlength{\tabcolsep}{6pt}
\resizebox{0.9\textwidth}{!}{
\begin{tabular}{l|ccc|ccc|ccc|ccc}
\toprule
\rowcolor{mygray}
\textbf{Method} 
& \multicolumn{3}{c|}{\textbf{1--3 ($\uparrow$)}} 
& \multicolumn{3}{c|}{\textbf{4--6 ($\uparrow$)}} 
& \multicolumn{3}{c|}{\textbf{7--9 ($\uparrow$)}} 
& \multicolumn{3}{c}{\textbf{ 9+ ($\uparrow$)}} \\
\rowcolor{mygray}
& OCA & AP & IR 
& OCA & AP & IR 
& OCA & AP & IR 
& OCA & AP & IR \\
\midrule
Show-o~\cite{xie2024show}                  & 64.6 & 79.6 & 0.262 & 55.4 & 65.2 & 0.194 & 47.8 & 37.1 & 0.163 & 27.6 & 20.9 & 0.014 \\
SEED-X~\cite{ge2024seed}                   & 68.3 & 53.1 & 0.407 & 47.7 & 29.8 & 0.401 & 26.2 & 12.7 & 0.322 & 22.9 & 9.7 & 0.245 \\
Emu-3~\cite{wang2024emu3}                  & 74.8 & 61.2 & 0.305 & 61.1 & 47.2 & 0.302 & 42.6 & 37.5 & 0.285 & 27.7 & 29.9 & 0.214 \\
LEDITS~\cite{tsaban2023ledits}             & 77.8 & 69.2 & 0.205 & 67.3 & 59.3 & 0.005 & 64.7 & 57.3 & -0.047 & 52.6 & 52.9 & -0.063 \\
LEDITS++~\cite{brack2024ledits++}          & 78.2 & 70.8 & 0.199 & 71.9 & 68.4 & -0.004 & 65.7 & 65.2 & -0.005 & 63.9 & 59.2 & -0.073 \\
LoMOE~\cite{chakrabarty2024lomoe}          & 88.6 & 80.6 & 0.236 & 72.1 & 74.4 & 0.203 & 68.0 & 63.7 & -0.004 & 60.5 & 52.9 & -0.055 \\
InstructPix2Pix~\cite{brooks2023instructpix2pix} & 88.2 & 88.1 & 0.322 & 87.5 & 83.0 & 0.293 & 79.4 & 72.9 & 0.183 & 74.9 & 69.1 & 0.091 \\
MagicBrush~\cite{zhang2023magicbrush}      & 89.1 & 87.8 & 0.305 & 87.4 & 83.7 & 0.296 & 81.7 & 72.5 & 0.254 & 78.3 & 71.2 & 0.154 \\
IC-Edit~\cite{zhang2025context}            & 89.3 & 88.1 & 0.402 & 85.2 & 84.8 & 0.383 & 83.2 & 79.4 & 0.298 & 78.8 & 74.3 & 0.294 \\
Any-Edit~\cite{yu2025anyedit}              & 89.4 & 88.4 & 0.418 & 86.9 & 85.0 & 0.395 & 83.8 & 80.1 & 0.345 & 79.1 & 76.9 & 0.302 \\
\hline
\rowcolor{blue!7} {IMAGHarmony (SD1.5)}    & {89.5} & {88.3} & {0.414} & {87.9} & {84.6} & {0.402} & {87.6} & \textbf{84.9} & {0.409} & {88.3} & {84.1} & \textbf{0.408} \\
\rowcolor{blue!7} \textbf{IMAGHarmony (SDXL)} & \textbf{89.9} & \textbf{88.5} & \textbf{0.419} & \textbf{88.4} & \textbf{85.2} & \textbf{0.404} & \textbf{87.7} & 84.8 & \textbf{0.411} & \textbf{89.2} & \textbf{85.1} & {0.399} \\
\bottomrule
\end{tabular}}

\end{table*}

\section{Experiments}\label{sec:exp}

\subsection{Implementation Details}

\noindent\textbf{Datasets.}
To address the lack of benchmarks for quantity-and-layout-consistent image editing, we construct HarmonyBench. The training split contains 200 image-caption pairs, uniformly distributed across 10 distinct object-count categories (ranging from 1 to 10+ instances) with 20 samples per category. Notably, this training process requires no explicit structural annotations such as dense masks. The evaluation split consists of 2K non-overlapping instruction-image pairs covering class, scene, and style editing. The source images are curated from diverse real-world datasets, including COCO~\cite{lin2015microsoftcococommonobjects} and OpenImages~\cite{kuznetsova_2020}, supplemented by high-quality synthesized images. As shown in Fig.~\ref{fig:example}, all annotations are rigorously verified through a two-stage pipeline: an initial coarse filtering using YOLO-World~\cite{cheng2024yoloworldrealtimeopenvocabularyobject}, followed by strict manual verification. To further assess cross-dataset generalization, we evaluate our method on the standard MagicBrush~\cite{zhang2023magicbrush} benchmark.

\noindent\textbf{Metrics.}
We evaluate QL-Edit using a comprehensive suite of objective and subjective metrics. For \textit{objective evaluation}, we report: (1) object count accuracy (OCA)~\cite{liu2019point}, which measures whether the generated image matches the target instance cardinality; (2) average precision (AP)~\cite{cho2024diagnostic}, which quantifies the spatial alignment between the generated objects and the reference layout; and (3) ImageReward (IR)~\cite{xu2023imagereward}, which reflects overall visual quality and text-to-image alignment. OCA and AP are computed using YOLO-World as the open-vocabulary detector. To validate this automatic pipeline, we compared YOLO-World predictions against manual annotations on a held-out subset, achieving over 95\% agreement. 
For \textit{subjective evaluation}, we report semantic correctness (SC) and oerall quality (OQ). We perform human verification on a randomly sampled subset of 500 test cases, where independent annotators evaluate whether the edited image satisfies the semantic instruction (SC) and rank the perceptual aesthetics (OQ). The human judgments align strongly with our automatic metrics, verifying the reliability of our evaluation protocol. All metrics are higher-is-better.

\noindent\textbf{Hyperparameters.}
All experiments are conducted on a single NVIDIA RTX 3090 GPU. We initialize the denoising UNet from the pretrained SDXL model~\cite{podell2023sdxlimprovinglatentdiffusion} and the cross-attention branch from the pretrained IP-Adapter~\cite{ye2023ip}. The HA module is configured with 8 attention heads. Input images are resized to $512 \times 512$, and a 5\% conditional dropout is applied during training to support classifier-free guidance (CFG)~\cite{ho2022classifier}. We optimize the trainable parameters in the HA module using AdamW with a constant learning rate of $2.5 \times 10^{-4}$, a batch size of 1, and 2K total training steps. During inference, we employ the DDIM sampler with 30 denoising steps. Following the notation defined in Section~\ref{sec:method}, the QL attention scaling factor $\alpha$, the perception cross-attention scale $\beta$, and the CFG scale $w$ are empirically set to 1.0, 1.0, and 5.0, respectively, unless otherwise specified.

\begin{figure*}[t]
  \centering
  \includegraphics[width=0.95\linewidth]{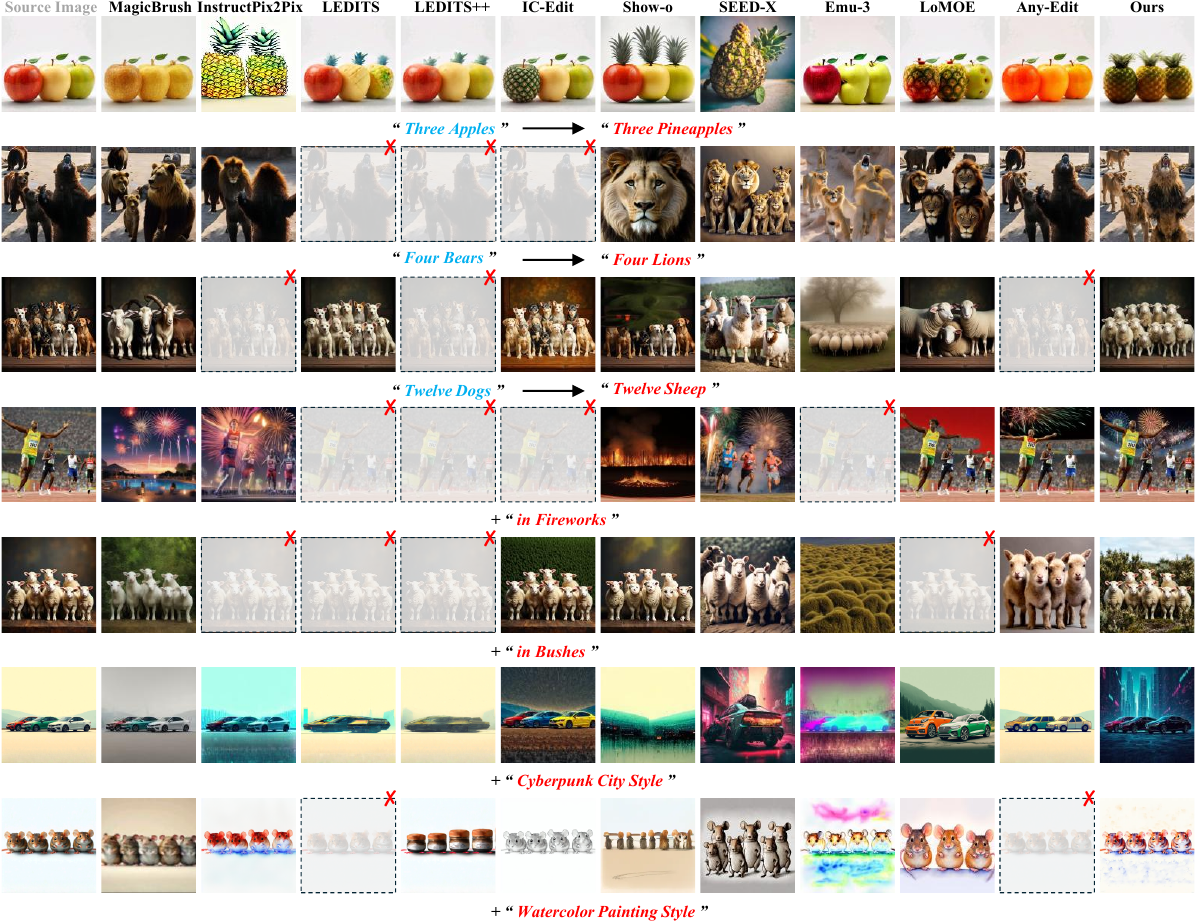}
  \caption{\textbf{Qualitative comparisons} with several state-of-the-art models on the HarmonyBench dataset. A red cross indicates an unsuccessful edit, where the output shows no difference compared to the source image. A white mask indicates that the corresponding image remains essentially unedited.}
  \label{fig:compare}
\end{figure*}

\subsection{Main Results}
We quantitatively compare the proposed IMAGHarmony with a broad range of state-of-the-art image editing methods across three categories: (1) mask-free editing approaches, including InstructPix2Pix~\cite{brooks2023instructpix2pix} and MagicBrush~\cite{zhang2023magicbrush}; (2) mask-based methods, such as LEDITS~\cite{tsaban2023ledits}, LEDITS++~\cite{brack2024ledits++}, LoMOE~\cite{chakrabarty2024lomoe}, IC-Edit~\cite{zhang2025context}, and Any-Edit~\cite{yu2025anyedit}; and (3) VLM-based approaches, including Show-o~\cite{xie2024show}, SEED-X~\cite{ge2024seed}, and Emu-3~\cite{wang2024emu3}.

\noindent\textbf{Quantitative Results.} 
Table~\ref{table:comparison} summarizes the performance on HarmonyBench across class, scene, and style editing tasks. IMAGHarmony consistently outperforms all SOTA methods across all metrics. 
Specifically: 
(1) Against the representative mask-free method, MagicBrush~\cite{zhang2023magicbrush}, we achieve substantial gains of +10.4 OCA and +6.5 AP in class editing, indicating superior object-level structural modeling. 
(2) Compared to the strongest baseline overall, Any-Edit~\cite{yu2025anyedit}, our SDXL framework still achieves a clear performance leap (+4.0 OCA and +3.3 AP on average) while preserving layout consistency, validating the effectiveness of the HA module. 
(3) Versus the unified VLM-guided model Show-o~\cite{xie2024show}, we surpass it by +13.8 OCA and +14.3 AP in scene editing, yielding significantly higher IR scores (0.217 vs. -0.134). This demonstrates our strong semantic alignment despite using a much more lightweight generative design.

We further analyze performance stratification by object count in Table~\ref{table:comparison_num}. 
Most existing methods degrade substantially as instance cardinality increases. For instance, MagicBrush~\cite{zhang2023magicbrush} drops from 89.1/87.8 (OCA/AP) in sparse scenes (1--3 objects) to 78.3/71.2 in dense scenes (9+ objects), while LEDITS++~\cite{brack2024ledits++} even yields negative IR (-0.073) under extreme dense conditions. 
In contrast, IMAGHarmony remains remarkably stable: the SDXL variant shows only a marginal reduction (e.g., OCA dropping slightly from 89.9 to 89.2) as visual complexity grows. 
These results across both SD1.5 and SDXL backbones confirm the exceptional robustness and generalization of our approach in diverse and challenging QL-Edit scenarios.

\begin{figure*}[t]
    \centering
    \includegraphics[width=0.95\linewidth]{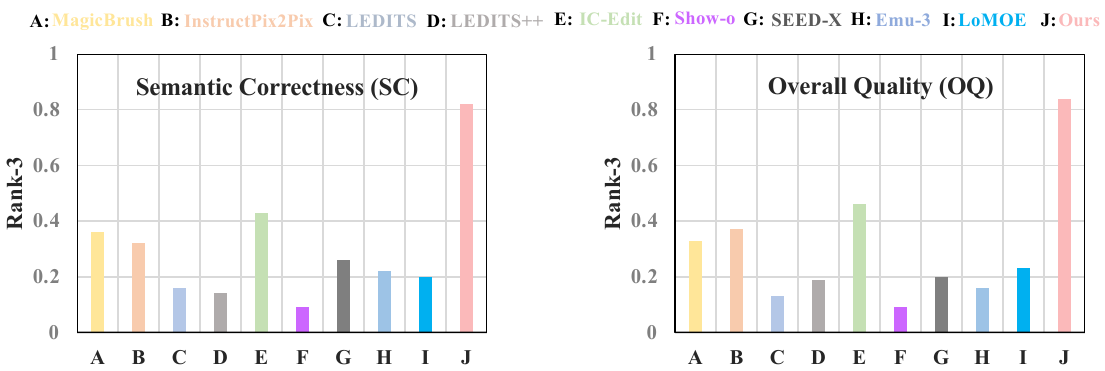}
    \caption{\textbf{User study results on HarmonyBench.} Our method receives the highest praise from users.}
    \label{fig:user_study}
    \vspace{-0.3cm}
\end{figure*}

\begin{table*}[t]
\centering
\caption{\textbf{Ablation study of the HA module and PNS strategy.} Green numbers indicate the absolute performance improvement compared to the Baseline.}
\label{tab:ablation}
\renewcommand{\arraystretch}{1.2} 
\newcommand{\gup}[1]{\textsuperscript{\textcolor{green!60!black}{\tiny (+#1)}}}

\begin{tabular}{l c c c c l l l}
\toprule
\multirow{2}{*}{\textbf{Settings}} & \multicolumn{2}{c}{\textbf{HA Module}} & \multicolumn{2}{c}{\textbf{PNS Strategy}} & \multirow{2}{*}{\textbf{OCA} ($\uparrow$)} & \multirow{2}{*}{\textbf{AP} ($\uparrow$)} & \multirow{2}{*}{\textbf{IR} ($\uparrow$)} \\
\cmidrule(lr){2-3} \cmidrule(lr){4-5}
 & \emph{QL Attn.} & \emph{Cross Attn.} & \emph{Train.} & \emph{Infer.} & & & \\
\midrule
Baseline & \textcolor{gray!40}{\xmark} & \textcolor{gray!40}{\xmark} & \textcolor{gray!40}{\xmark} & \textcolor{gray!40}{\xmark} & 43.2 & 21.7 & 0.161 \\
\midrule
B1       & \textcolor{gray!40}{\xmark} & \cmark & \textcolor{gray!40}{\xmark} & \textcolor{gray!40}{\xmark} & 48.6\gup{5.4} & 56.9\gup{35.2} & 0.193\gup{0.032} \\
B2       & \cmark & \cmark & \textcolor{gray!40}{\xmark} & \textcolor{gray!40}{\xmark} & 84.6\gup{41.4} & 82.4\gup{60.7} & 0.366\gup{0.205} \\
B3       & \cmark & \cmark & \textcolor{gray!40}{\xmark} & \cmark & 85.1\gup{41.9} & 82.7\gup{61.0} & 0.369\gup{0.208} \\
B4       & \cmark & \cmark & \cmark & \textcolor{gray!40}{\xmark} & 86.7\gup{43.5} & 84.4\gup{62.7} & 0.391\gup{0.230} \\
\midrule
\rowcolor{blue!5} \textbf{Ours} & \cmark & \cmark & \cmark & \cmark & \textbf{88.8}\gup{45.6} & \textbf{85.9}\gup{64.2} & \textbf{0.408}\gup{0.247} \\
\bottomrule
\end{tabular}
\end{table*}

\noindent\textbf{Qualitative Results.}
Fig.~\ref{fig:compare} provides visual comparisons across diverse QL-Edit scenarios. 
(1) For class editing, IMAGHarmony precisely transforms object categories (\emph{e.g., ``Apples'' $\to$ ``Pineapples''}) while preserving the original instance count and spatial layout. In contrast, MagicBrush~\cite{zhang2023magicbrush} frequently fails to execute the semantic shift, whereas other methods suffer from severe layout disruption and count drift under high-cardinality queries (\emph{e.g., ``Seven Bears''}). 
(2) In scene editing, our method seamlessly modifies background contexts (\emph{e.g., +``Fireworks''}) while leaving the foreground instances completely intact. Competitors such as LEDITS++~\cite{brack2024ledits++} and IC-Edit~\cite{zhang2025context} exhibit severe foreground-background entanglement, while SEED-X~\cite{ge2024seed} struggles to maintain quantity consistency. 
(3) In style editing, IMAGHarmony successfully imparts global stylistic transformations (\emph{e.g., +``Cyberpunk''}) without compromising structural integrity or original semantics. Conversely, existing methods frequently over-stylize and distort the underlying geometry; for instance, Show-o~\cite{xie2024show} mistakenly alters the intrinsic object categories.  Overall, the proposed IMAGHarmony achieves superior disentanglement between editing instructions and spatial priors, delivering highly coherent and structurally consistent edits even in complex multi-object scenes.

\noindent\textbf{User Study.}
To complement our objective metrics, we conducted a user study with 40 participants to evaluate IMAGHarmony from a human-centered perspective. We randomly sampled 50 editing instructions covering all three QL-Edit tasks (class, scene, and style editing) and generated the corresponding results using our method and the leading baselines. For each prompt, participants were asked to select the top-3 outputs based on two primary criteria: (1) Semantic Correctness (SC)~\cite{zhang2023magicbrush}, which evaluates the accuracy of the semantic modification while assessing object count and layout preservation; and (2) Overall Quality (OQ)~\cite{brooks2023instructpix2pix}, which assesses perceptual realism, visual coherence, and aesthetic appeal. To mitigate evaluation bias, all generated images were presented in a randomized and anonymized order, with each participant evaluating a distinct subset of prompt-image pairs to ensure balanced coverage. 
As illustrated in Fig.~\ref{fig:user_study}, IMAGHarmony consistently achieves the highest preference rates on both SC and OQ. Notably, our method outperforms the second-best baseline, IC-Edit, by a substantial margin, securing nearly double the preference rate across both metrics. This indicates that our framework not only provides superior semantic alignment and structural preservation, but also delivers greater perceptual realism. These subjective findings strongly corroborate our quantitative evaluations, validating the practical effectiveness of IMAGHarmony in real-world multi-object editing scenarios.

\subsection{Ablation and Analysis}

\noindent\textbf{Effectiveness of the Proposed Modules.}
To evaluate the individual contributions of the harmony-aware (HA) module and the preference-guided noise selection (PNS) strategy, we conduct a detailed ablation study across five variants (summarized in Table~\ref{tab:ablation}):
\begin{itemize}
\item \textbf{Baseline}: No HA module and no PNS strategy; source-image features are injected via a standard cross-attention layer.
\item \textbf{B1}: Uses only the cross-attention layer of the HA module (omitting QL attention); no PNS strategy.
\item \textbf{B2}: Uses the full HA module; no PNS strategy.
\item \textbf{B3}: Uses the full HA module; applies the PNS strategy \emph{only} at inference.
\item \textbf{B4}: Uses the full HA module; applies the PNS strategy \emph{only} during training.
\end{itemize}

\begin{figure}[t]
    \centering
\includegraphics[width=0.9\linewidth]{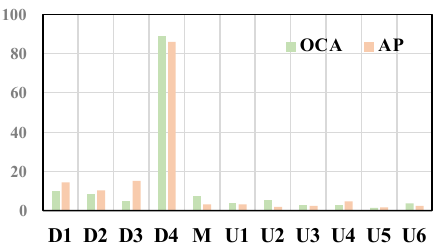}
    \caption{\textbf{Impact of injection locations within the U-Net.} Feature injection at the lowest-resolution bottleneck (D4) achieves the peak performance in both instance cardinality (OCA) and spatial alignment (AP), highlighting its critical role in structural control.}
    \label{fig:diff_block}
    \vspace{-0.3cm}
\end{figure}

As shown in Table~\ref{tab:ablation}, the full IMAGHarmony pipeline achieves the best results across all metrics (88.8 OCA, 85.9 AP, and 0.408 IR). 
First, we analyze the effectiveness of the HA module. The naive Baseline performs poorly across all metrics (OCA 43.2 / AP 21.7 / IR 0.161). Introducing only the cross-modal interaction (B1) markedly improves spatial alignment (+35.2 AP) but yields limited gains in instance counting (+5.4 OCA). However, when the complete HA module (incorporating QL attention, B2) is enabled, performance surges dramatically to 84.6 OCA and 82.4 AP. This demonstrates that explicit quantity modeling paired with implicit layout reasoning is necessary for stable structural control.
Next, we decouple the impact of the PNS strategy into its training and inference stages. Applying PNS solely during inference (B3) provides marginal benefits over B2 (e.g., OCA +0.5). Conversely, introducing PNS during the training phase (B4) leads to more substantial improvements (+2.1 OCA and +2.0 AP over B2). This indicates that exposing the model to preferred noise trajectories during optimization is highly critical for trajectory stability and structural fidelity.
Finally, the complete framework leverages the PNS strategy in both training and inference alongside the HA module. Notably, AP and IR continue to rise consistently from B2 through B4 to our final model. This synergistic effect, consistent with our qualitative visualizations, proves that the dual-stage PNS strategy perfectly complements the HA module, significantly strengthening both layout consistency and overall perceptual image quality.

\begin{table}[t]
\centering
\caption{\textbf{Quantitative results of different fusion strategies} on the HarmonyBench dataset.}
\label{tab:fusion}
\begin{tabular}{l|ccc}
\toprule
\rowcolor{mygray}\textbf{Methods} & \textbf{OCA ($\uparrow$)} & \textbf{AP ($\uparrow$)} & \textbf{IR ($\uparrow$)} \\
\midrule
Q-Former   & 42.3 & 41.9 & 0.134 \\
MLP  & 44.6 & 36.8 & 0.078 \\
Gated Attention  & 67.4 & 55.9 & 0.194 \\
\hline
\rowcolor{blue!7} \textbf{Ours} & \textbf{88.8} & \textbf{85.9} & \textbf{0.408} \\
\bottomrule
\end{tabular}
\vspace{-0.3cm}
\end{table}

\begin{figure}[t]
    \centering
    \includegraphics[width=0.95\linewidth]{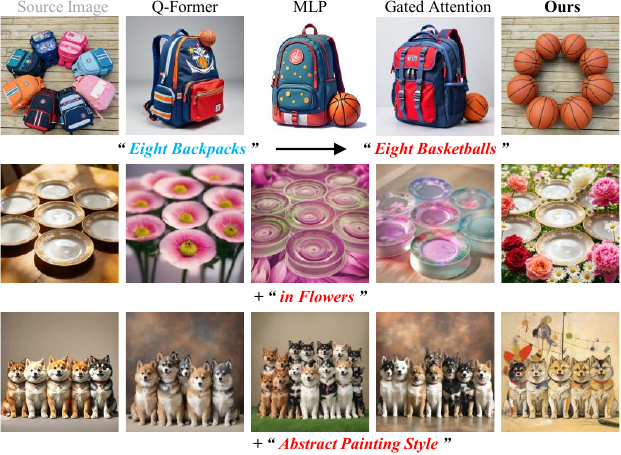}
\caption{\textbf{Qualitative comparison of feature fusion strategies.} Compared to Q-Former, MLP, and Gated Attention, our HA module consistently achieves superior instance cardinality and layout preservation across class, scene, and style editing, while baselines often suffer from count errors and structural distortions.}
    \label{fig:fusion}
    \vspace{-0.3cm}
\end{figure}

\noindent\textbf{Injection Location Analysis.}
As illustrated in Fig.~\ref{fig:diff_block}, we evaluate the impact of injecting the harmony features at various network depths within the UNet, spanning the downsampling (D1--D4), middle (M), and upsampling (U1--U6) blocks. The results demonstrate a clear performance peak at the lowest-resolution bottleneck (D4), which significantly outperforms all other insertion configurations in both object count accuracy (OCA) and average precision (AP). 
This empirical finding perfectly corroborates Proposition 1 established in our theoretical analysis (Section~\ref{sec:theory_down4}). Due to the hierarchical spectral bias of the U-Net architecture, injecting spatial and quantitative priors at shallower layers (\emph{e.g.}, D1 or D2) predominantly affects high-frequency texture details rather than global geometry, resulting in inferior structural control. By introducing the HA module precisely at the D4 bottleneck, we explicitly modulate the low-frequency components. This strategic placement allows the model to effectively govern macroscopic spatial arrangements and instance cardinality, thereby maximizing structural consistency. Moreover, operating at this highly compressed spatial resolution introduces negligible computational overhead, ensuring that both training and inference latency remain virtually unchanged.

\begin{figure}[t]
    \centering
    \includegraphics[width=0.95\linewidth]{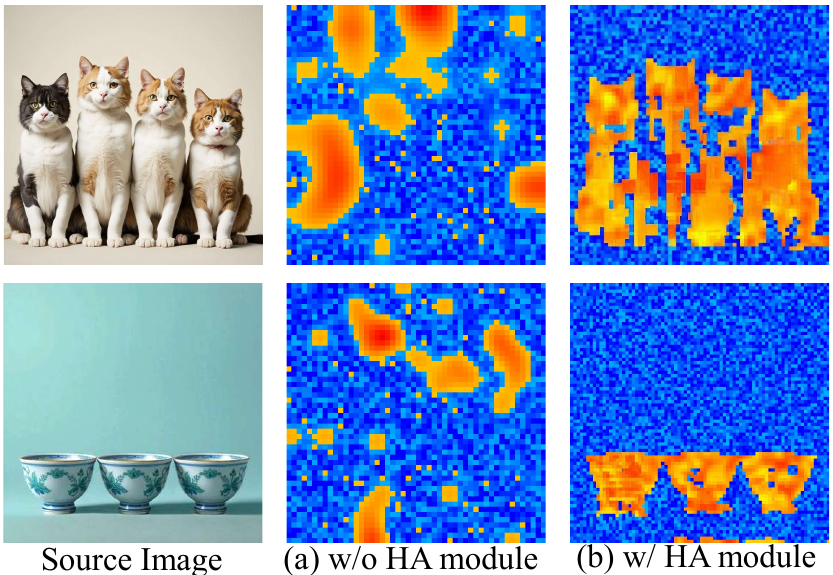}
\caption{\textbf{Visualization of internal attention maps.} (a) The baseline exhibits diffuse activation patterns. (b) Our HA module produces concentrated attention peaks aligned with target instances, demonstrating superior modeling of instance cardinality and spatial arrangements.}
    \label{fig:eff_feature}
    \vspace{-0.5cm}
\end{figure}

\noindent\textbf{Feature Fusion Strategies.}  
Building upon the ablation of the HA module, which demonstrates the necessity of explicit quantity-layout priors for structural consistency, we further compare our QL attention against three representative feature-fusion strategies: Q-Former~\cite{li2023blip}, MLP fusion~\cite{fukui2016multimodal}, and Gated Attention~\cite{zhang2021vinvl}. 
As illustrated in Fig.~\ref{fig:fusion}, we observe distinct behaviors across different editing scenarios: 
(1) For class editing (top row), our HA module consistently replaces the object category (\emph{e.g., ``Eight Backpacks'' $\to$ ``Eight Basketballs''}) while faithfully preserving the exact object count and spatial layout. Although the three competing fusion strategies achieve semantic category transfer, their insufficient structural modeling leads to severe layout distortions and count errors in multi-object scenes (\emph{e.g.}, MLP fusion incorrectly yields five basketballs). 
(2) For scene editing (middle row), our method successfully introduces new scene semantics (\emph{e.g., +``In Flowers''}) while maintaining object integrity and positional consistency. In contrast, MLP and Gated Attention tend to severely entangle foreground and background semantics, and Q-Former still exhibits count omissions, resulting in incomplete edits. 
(3) For style editing (bottom row), HA flawlessly transfers the target visual style (\emph{e.g., +``Abstract Painting Style''}) without altering object semantics or spatial arrangements. Other methods not only incur count errors but also produce semantic distortions and visual artifacts.
Consistent with this qualitative evidence, the quantitative results in Table~\ref{tab:fusion} further corroborate the superiority of the HA module: our proposed QL attention significantly surpasses all three representative fusion strategies (Q-Former, MLP, and Gated Attention) across every metric (OCA, AP, and IR). 

\begin{figure}[t]
    \centering
    \includegraphics[width=0.95\linewidth]{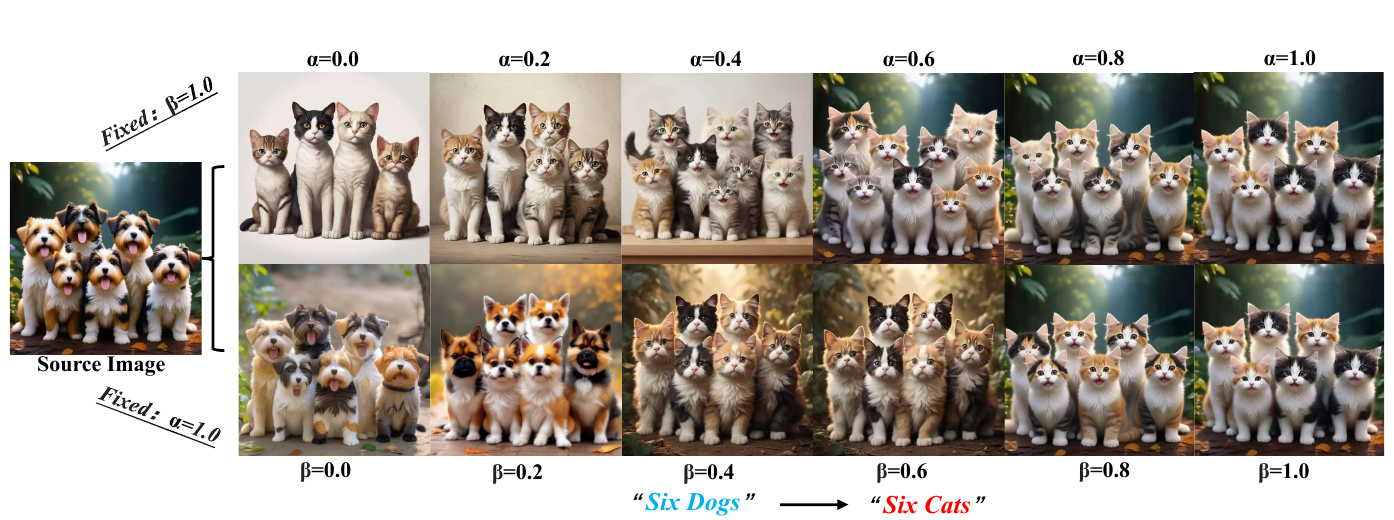}
    \caption{\textbf{Effect of attention weights on editing quality.} Setting both $\alpha$ and $\beta$ to 1.0 produces the best results.}
    \label{fig:super_params}
    \vspace{-0.5cm}
\end{figure}

\begin{table}[t]
\centering
\caption{\textbf{Ablation study of VLM evaluators and key hyperparameters.}}
\label{tab:unified_abl}
\renewcommand{\arraystretch}{1.15}
\begin{tabular}{l| c c c c}
\toprule
\rowcolor{mygray}\textbf{Factors} & \textbf{Types} & \textbf{OCA $\uparrow$} & \textbf{AP $\uparrow$} & \textbf{IR $\uparrow$} \\
\midrule
\multicolumn{5}{l}{\textit{\textcolor{gray}{(1) VLM Evaluator}}} \\
\textbf{\multirow{2}{*}{VLM}} & GPT-4V & 88.7    & 85.8  & 0.407       \\
             & CogVLM2    & \cellcolor{blue!7}\textbf{88.8} & \cellcolor{blue!7}\textbf{85.9} & \cellcolor{blue!7}\textbf{0.408} \\
\cdashline{1-5}[.6pt/2pt] 
\multicolumn{5}{l}{\textit{\textcolor{gray}{(2) Effect of Candidate Seeds}}} \\
\textbf{\multirow{3}{*}{K}} & 3           & 87.4  & 83.9      & 0.386    \\
                     & 5           & \textbf{88.8}  & 85.9    & 0.408    \\
    & 10         & \cellcolor{blue!7}\textbf{88.8} & \cellcolor{blue!7}\textbf{86.0} & \cellcolor{blue!7}\textbf{0.411} \\
\cdashline{1-5}[.6pt/2pt] 
\multicolumn{5}{l}{\textit{\textcolor{gray}{(3) Shallow Denoising Steps}}} \\
\textbf{\multirow{3}{*}{Steps}} & 10       & 88.8  & 85.9    & 0.408      \\
                       & 20       & 88.8  & 85.9    & 0.412       \\
      & 30       & \cellcolor{blue!7}\textbf{89.8} & \cellcolor{blue!7}\textbf{86.2} & \cellcolor{blue!7}\textbf{0.414} \\
\bottomrule
\end{tabular}
\vspace{-0.4cm}
\end{table}

\begin{figure*}[t]
\centering
\includegraphics[width=0.98\linewidth]{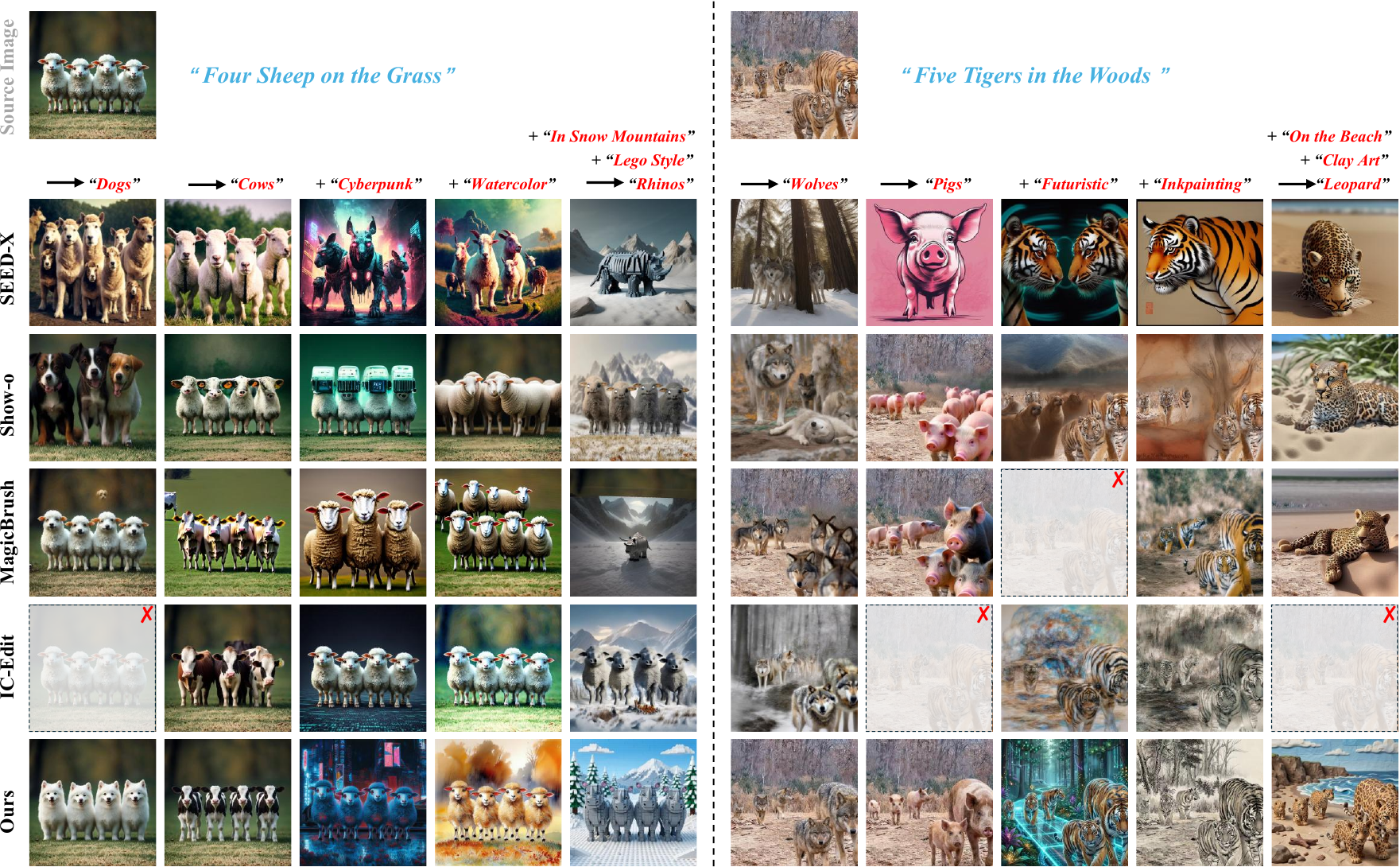}
     \caption{\textbf{Qualitative results with a fixed source image.}  IMAGHarmony produces more stable and consistent images. A white mask indicates that the corresponding image remains essentially unedited.
     }
     \label{fig:com1}
 \end{figure*}

\noindent\textbf{Visualization of Internal Feature Maps.}
To explicitly validate the effectiveness of the HA module in modeling object quantity and spatial layout, we visualize the internal attention distributions of the diffusion model. As shown in Fig.~\ref{fig:eff_feature}, the heatmaps are generated by averaging the activation maps from the ``Down4'' blocks of the UNet. 
Given a source image with multiple distinct objects, the baseline model (without the HA module) exhibits a highly dispersed and diffuse activation pattern (Fig.~\ref{fig:eff_feature}a). This weak structural focus fails to isolate individual instances, inherently leading to the missing or misaligned objects frequently observed in baseline generations. In stark contrast, integrating the HA module (Fig.~\ref{fig:eff_feature}b) produces highly concentrated, sharp attention peaks that perfectly align with the spatial centroids of the target objects. This compelling visual evidence confirms that the HA module successfully injects robust quantity and layout priors into the diffusion backbone, thereby enforcing structural consistency.

\begin{figure}[t]
    \centering
    \includegraphics[width=0.95\linewidth]{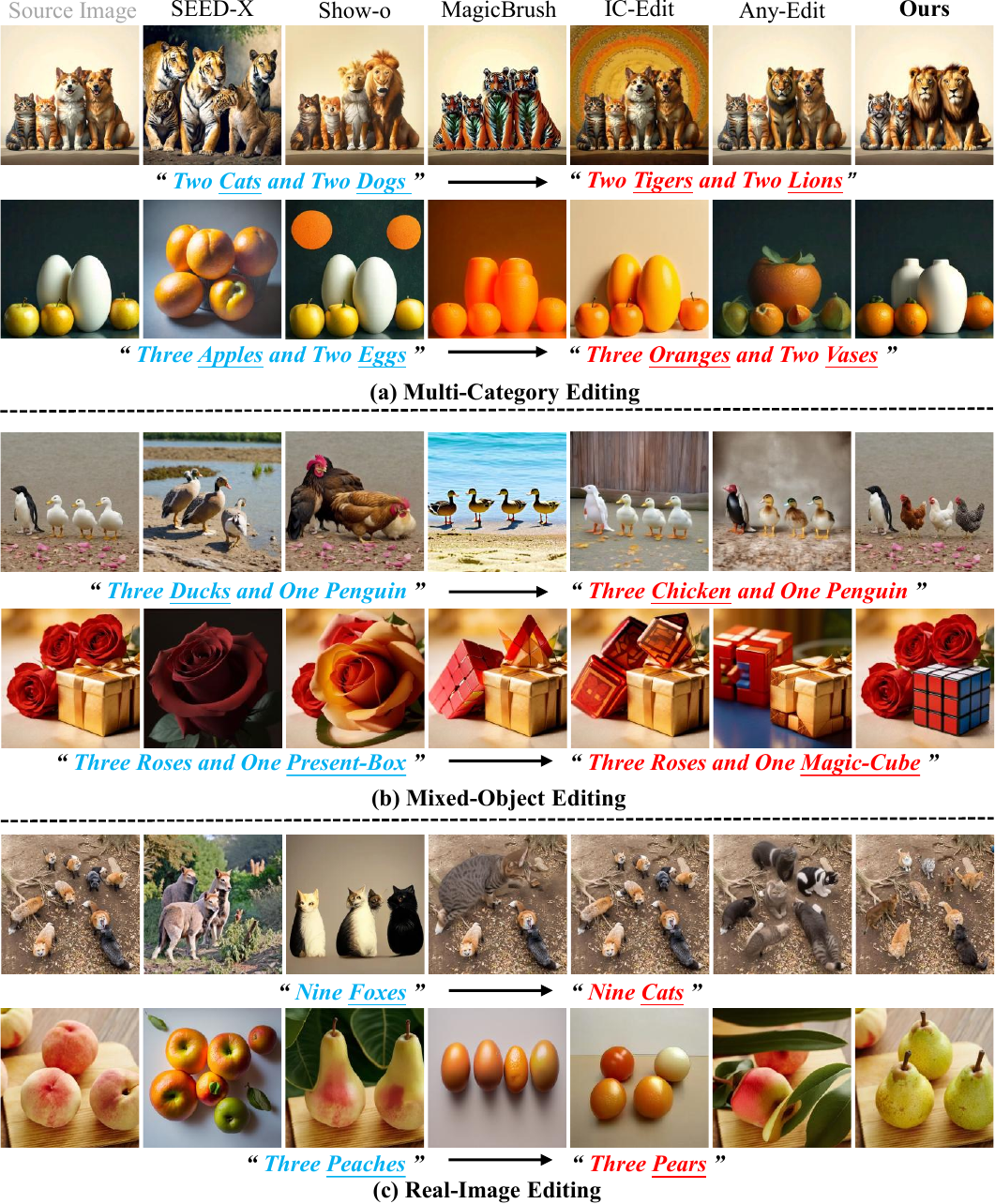}
\caption{\textbf{Qualitative comparison in complex editing scenarios.} We evaluate IMAGHarmony across three settings: \textbf{(a) Multi-Category Editing}, transforming multiple distinct classes simultaneously; \textbf{(b) Mixed-Object Editing}, modifying target categories while anchoring non-target instances; and \textbf{(c) Real-Image Editing}, showcasing robust zero-shot generalization to cluttered real-world photographs.}
    \label{fig:more_results}
    \vspace{-0.3cm}
\end{figure}

\noindent\textbf{Impact of Hyperparameters.}
We further investigate the impact of two key hyperparameters, $\alpha$ and $\beta$, which respectively control the influence of the HA module's structural features and the perception cross-attention branch. As illustrated in Fig.~\ref{fig:super_params}, the optimal balance of editing quality is achieved when both $\alpha=1.0$ and $\beta=1.0$. 
Progressively increasing $\alpha$ heightens the model's sensitivity to instance cardinality and spatial arrangements, whereas increasing $\beta$ strengthens its fidelity to the target semantic category. This empirical trend perfectly aligns with our architectural design: the HA module explicitly injects robust quantity-layout priors, while the dedicated cross-attention branch ensures rigorous semantic alignment.

\noindent\textbf{Analysis of Additional Variants.}
Table~\ref{tab:unified_abl} reports the effects of three practical operational factors in our framework. 
First, employing GPT-4V~\cite{achiam2023gpt} versus CogVLM2~\cite{hong2024cogvlm2} as the seed-scoring evaluator yields nearly identical quantitative results. This indicates that the PNS strategy is highly robust and generalizes well, remaining fundamentally agnostic to the specific choice of the underlying VLM. 
Second, enlarging the candidate pool size \(K\) brings initial modest gains but quickly saturates beyond \(K=5\). Given that the inference overhead scales roughly linearly with \(K\), we select \(K=5\) as an optimal trade-off between structural fidelity and computational efficiency. 
Finally, increasing the number of shallow denoising steps provides only marginal structural improvements at the expense of prolonged runtime. Consequently, we adopt 10 steps as a balanced and efficient configuration for practical deployment.

\begin{table}[t]
\centering
\small
\setlength{\tabcolsep}{6pt}
\caption{\textbf{Quantitative comparison on the MagicBrush dataset.} IMAGHarmony achieves state-of-the-art performance across all metrics, demonstrating superior generalization to real-world editing tasks.}
\label{tab:MagicBrush}
\begin{tabular}{l|ccc}
\toprule
\rowcolor{mygray} \textbf{Methods} & \textbf{OCA ($\uparrow$)} & \textbf{AP ($\uparrow$)} & \textbf{IR ($\uparrow$)} \\
\midrule
SEED-X          & 40.8 & 21.4 & 0.337 \\
Show-o          & 51.2 & 51.9 & 0.114 \\
Emu-3           & 58.2 & 47.0 & 0.187 \\
LEDITS          & 67.2 & 63.9 & 0.017 \\
LEDITS++        & 74.8 & 72.5 & 0.093 \\
LoMOE           & 82.3 & 78.3 & 0.105 \\
Any-Edit        & 84.3 & 80.2 & 0.335 \\
InstructPix2Pix & 85.4 & 78.7 & 0.314 \\
MagicBrush      & \underline{88.2} & 77.1 & 0.345 \\
IC-Edit         & 88.6 & \underline{84.8} & \underline{0.436} \\
\midrule
\rowcolor{blue!7} \textbf{Ours} & \textbf{91.7} & \textbf{89.1} & \textbf{0.479} \\
\bottomrule
\end{tabular}
\vspace{-0.3cm}
\end{table}

 \begin{table*}[t]
\centering
\caption{\textbf{Efficiency comparison regarding data scale, model parameters, and inference overhead.} Our method achieves superior performance with orders of magnitude less training data and fewer trainable parameters while maintaining the fastest inference speed. (Lower is better for all metrics.)}
\label{tab:inference}
\renewcommand{\arraystretch}{1.05}
\setlength{\tabcolsep}{7pt}
\begin{tabular}{l|ccccc >{\columncolor{blue!7}}c} 
\toprule
\rowcolor{mygray}
\textbf{Metrics} & \textbf{SEED-X} & \textbf{Show-o} & \textbf{MagicBrush} & \textbf{IC-Edit} & \textbf{Any-Edit} & \cellcolor{mygray}\textbf{Ours} \\
\midrule
\textbf{Training Data ($\downarrow$)}   & 158.0M & 2.0B & 8.8K & 50.0K & 2.5M & \textbf{0.2K} \\
\textbf{Train. Params ($\downarrow$)}   & 2.6B   & 1.3B & 1.1B & 214.0M & 1.8B & \textbf{10.6M} \\
\textbf{Speed ($\downarrow$)}           & 12.4s  & 10.4s & 24.2s & 9.7s & 11.5s & \textbf{6.6s} \\
\textbf{Model Scale ($\downarrow$)}     & 13B    & 1.3B & 1.1B & 12B & 1.8B & \textbf{0.78B} \\
\bottomrule
\end{tabular}
\end{table*}

\subsection{Additional Qualitative Results}

\noindent\textbf{Results with a Fixed Source Image.}
Fig.~\ref{fig:com1} illustrates both class and style editing applied to a fixed source image. 
In class editing (\emph{e.g., ``Sheep'' $\to$ ``Dogs''}), IMAGHarmony precisely executes semantic transitions while preserving the original object count and coarse layout. In contrast, SOTA methods~\cite{ge2024seed,xie2024show,zhang2023magicbrush,zhang2025context} frequently suffer from incorrect category replacement or severe cardinality drift. 
Similarly, for style editing (\emph{e.g., +``Cyberpunk''}), our method successfully harmonizes the target aesthetic while maintaining structural integrity. Conversely, existing methods often suffer from structural collapse under strong stylistic transformations; for instance, IC-Edit fails to preserve object geometry when conditioned on prompts like ``Futuristic''.

\noindent\textbf{Multi-Category Editing.} 
Fig.~\ref{fig:more_results}(a) presents the challenging scenario of simultaneously transforming multiple distinct categories (\emph{e.g., ``Two Cats and Two Dogs'' $\to$ ``Two Tigers and Two Lions''}). VLM-based approaches~\cite{ge2024seed,xie2024show} and MagicBrush~\cite{zhang2023magicbrush} suffer from severe semantic entanglement or overfit to a single category. While IC-Edit~\cite{zhang2025context} and Any-Edit~\cite{yu2025anyedit} better retain layouts, they still introduce semantic bleeding and local distortions. In contrast, IMAGHarmony seamlessly transforms both targets while preserving their respective instance counts and spatial arrangements.

\noindent\textbf{Mixed-Object Editing.} 
Fig.~\ref{fig:more_results}(b) evaluates localized editing, where target objects are modified while non-target instances must remain untouched. Early baselines exhibit severe over-editing, unintentionally altering the entire scene. Recent methods like IC-Edit~\cite{zhang2025context} and Any-Edit~\cite{yu2025anyedit} successfully retain non-target categories but struggle to anchor their precise details and poses (\emph{e.g.}, distorted chickens or shifted penguins). In contrast, IMAGHarmony precisely edits target objects while keeping the rest of the scene unchanged.

\noindent\textbf{Real-Image Editing.} 
To assess zero-shot generalization, Fig.~\ref{fig:more_results}(c) shows editing results on in-the-wild photographs (COCO, OpenImages) featuring severe occlusions and cluttered backgrounds. Under such complex visual distributions, state-of-the-art models (\emph{e.g.}, IC-Edit~\cite{zhang2025context}, Any-Edit~\cite{yu2025anyedit}) frequently corrupt object quantity, details, or underlying scene geometry (\emph{e.g.}, hallucinating extra fruits or blending semantics). Remarkably, IMAGHarmony robustly generates highly realistic target categories while consistently anchoring the original instance cardinality and spatial layouts.

\subsection{Generalization and Efficiency}
\noindent\textbf{Generalization on the MagicBrush Dataset.} 
To assess the zero-shot generalization capability of our approach on diverse, unseen instruction distributions, we evaluate IMAGHarmony on the widely adopted MagicBrush~\cite{zhang2023magicbrush} benchmark. We compare against ten recent state-of-the-art baselines, encompassing VLM-guided models, mask-based frameworks, and mask-free instruction-driven methods. 
As detailed in Table~\ref{tab:MagicBrush}, IMAGHarmony establishes a new state-of-the-art across all metrics, achieving 91.7 OCA, 89.1 AP, and 0.479 IR. Notably, compared to the strongest competing baseline, IC-Edit~\cite{zhang2025context}, our method yields substantial absolute gains of +3.1 in OCA, +4.3 in AP, and +0.043 in IR. It is worth emphasizing that while top-performing competitors like IC-Edit and Any-Edit heavily rely on explicit mask-based localized control, IMAGHarmony achieves superior spatial alignment and instance preservation in a fully mask-free manner. These results confirm that the structural priors learned by our framework generalize robustly to real-world image editing tasks.

\begin{figure}[t]
    \centering
    \includegraphics[width=0.98\linewidth]{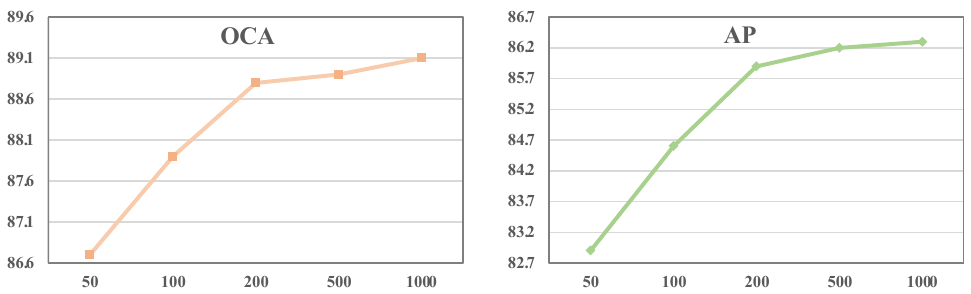}
\caption{\textbf{Training data scaling analysis.} The performance of IMAGHarmony saturates at approximately 200 samples, underscoring the high data efficiency of our proposed framework.}
    \label{fig:data_scaling}
    \vspace{-0.3cm}
\end{figure}

\noindent\textbf{Data and Parameter Efficiency.} 
Table~\ref{tab:inference} compares IMAGHarmony against five recent baselines across training data, trainable parameters, inference speed, and overall model scale under identical hardware settings. IMAGHarmony demonstrates overwhelming superiority across all efficiency metrics. 
Specifically, our framework requires merely 200 training images (0.2K), which is approximately $250\times$ fewer than IC-Edit~\cite{zhang2025context} (50.0K) and orders of magnitude fewer than VLM-guided models like SEED-X~\cite{ge2024seed} (158.0M). Furthermore, the HA module introduces only 10.6M trainable parameters, roughly $20\times$ smaller than IC-Edit (214.0M) and vastly more lightweight than Any-Edit~\cite{yu2025anyedit} (1.8B). During inference, our method achieves the fastest generation speed (6.6s), running 32\% faster than IC-Edit (9.7s) and nearly $4\times$ faster than MagicBrush~\cite{zhang2023magicbrush} (24.2s). Coupled with the most compact overall model scale (0.78B), these results demonstrate that IMAGHarmony achieves superior editing performance with only a fraction of the computational and data costs of existing frameworks.

\noindent\textbf{Data Scaling.}
A key concern is whether 200 training samples are sufficient for generalization. We trained IMAGHarmony on subsets of varying sizes: \{50, 100, 200, 500, 1000\} images.
As illustrated in Fig.~\ref{fig:data_scaling}, the OCA/AP improves rapidly from 50 to 100 samples and saturates around 200. Increasing the data size to 1000 yields marginal improvements but significantly increases training time.
This confirms that our HA module learns the generalized representation of "quantity and layout" efficiently, leveraging the strong semantic prior of the frozen SDXL backbone.

\noindent\textbf{Backbone Generalizability.}
To verify the architectural agnosticism of IMAGHarmony, we evaluate its plug-and-play capability across diverse diffusion backbones, ranging from traditional U-Net structures (SD1.5~\cite{rombach2022high} and SDXL~\cite{podell2023sdxlimprovinglatentdiffusion}) to the state-of-the-art DiT architecture (FLUX.1-dev~\cite{flux2024}).
As reported in Table~\ref{tab:backbone_scalability}, our framework consistently yields massive performance leaps across all generative engines. Notably, when integrated into the high-capacity FLUX.1-dev, IMAGHarmony drastically boosts OCA and AP from 49.8 and 28.6 to 90.4 and 87.9, respectively, while significantly enhancing image-text alignment (IR from 0.196 to 0.472). This demonstrates that our HA module and PNS strategy effectively resolve multi-object coordination bottlenecks inherent to both U-Net and DiT models. Such robust scalability underscores IMAGHarmony as a universal, lightweight enhancement for next-generation large-scale foundation models.

\begin{table}[t]
\centering
\small
\caption{\textbf{Generalizability across diffusion backbones.} IMAGHarmony consistently delivers massive performance gains across both traditional U-Net (SD1.5, SDXL) and state-of-the-art DiT (FLUX.1-dev) architectures.}
\label{tab:backbone_scalability}
\renewcommand{\arraystretch}{1.1}
\setlength{\tabcolsep}{8pt}
\begin{tabular}{l|ccc}
\toprule
\rowcolor{mygray} \textbf{Backbone} & \textbf{OCA ($\uparrow$)} & \textbf{AP ($\uparrow$)} & \textbf{IR ($\uparrow$)} \\
\midrule
SD1.5           & 32.4 & 18.7 & 0.078 \\
\rowcolor{blue!7} \quad \textbf{+ Ours} & \textbf{88.3} & \textbf{85.5} & \textbf{0.401} \\
\midrule
SDXL            & 43.2 & 21.7 & 0.161 \\
\rowcolor{blue!7} \quad \textbf{+ Ours} & \textbf{88.8} & \textbf{85.9} & \textbf{0.408} \\
\midrule
FLUX.1-dev      & 49.8 & 28.6 & 0.196 \\
\rowcolor{blue!7} \quad \textbf{+ Ours} & \textbf{90.4} & \textbf{87.9} & \textbf{0.472} \\
\bottomrule
\end{tabular}
\vspace{-0.3cm}
\end{table}

\subsection{Limitations}
\begin{figure}[t]
    \centering
    \includegraphics[width=0.95\linewidth]{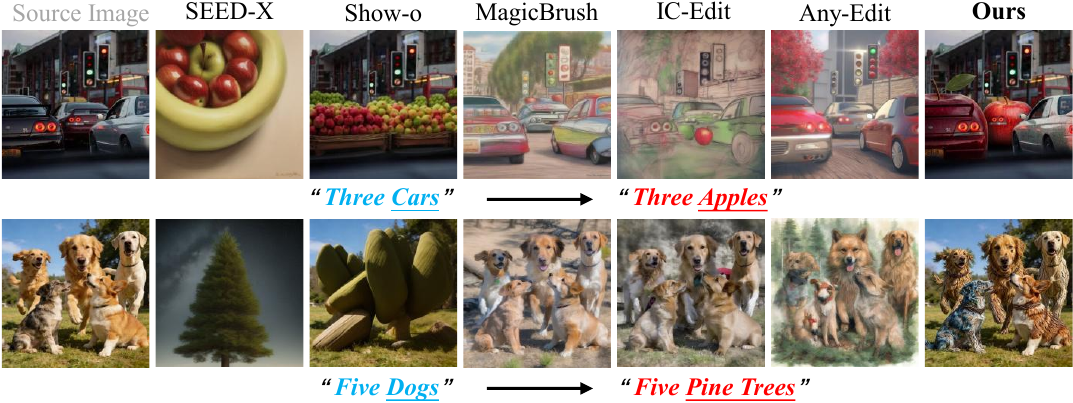}
    \caption{\textbf{Failure cases in scenarios with extreme geometric discrepancies.} While our framework preserves instance layouts, this rigid constraint can hinder natural shape transformation when the source and target geometries are fundamentally incompatible (\emph{e.g., ``Cars'' $\to$ ``Apples''}), leading to unnatural hybrid artifacts.}
    \label{fig:limitation}
    \vspace{-0.5cm}
\end{figure}

While IMAGHarmony excels at preserving instance cardinality and spatial arrangements, its strict adherence to structural priors becomes a double-edged sword when the editing instruction necessitates a radical transformation of intrinsic geometry. 
As illustrated in Fig.~\ref{fig:limitation}, when the source and target categories possess fundamentally incompatible shapes (\emph{e.g., ``Three Cars'' $\to$ ``Three Apples''} or \emph{``Five Dogs'' $\to$ ``Five Pine Trees''}), the model's rigid layout retention prevents natural topological morphing. Consequently, instead of restructuring the global boundaries, the diffusion backbone tends to merely project target textures onto the source silhouettes, resulting in unnatural hybrid artifacts (\emph{e.g.}, apple-textured vehicles with taillights or dog-shaped trees). In these extreme cases, while baseline methods completely lose the object count or hallucinate entirely new scenes, IMAGHarmony over-constrains the spatial layout. 
Future research will explore adaptive structural relaxation, allowing the model to dynamically loosen spatial constraints when severe geometric conflicts are detected, and integrating geometry-aware morphing mechanisms to better balance layout fidelity with natural semantic deformation.

\section{Conclusion}\label{sec:con}
In this paper, we explored quantity-and-layout-consistent image editing (QL-Edit) and proposed IMAGHarmony, a plug-and-play framework for controllable multi-object editing. The core of our approach is the harmony-aware (HA) module, which seamlessly fuses perception semantics to explicitly model object counts and implicitly capture spatial relations. To further enhance structural consistency, we introduced a preference-guided noise selection (PNS) strategy that identifies the most semantically aligned initial noise for generation. Furthermore, to support rigorous evaluation, we constructed HarmonyBench, a comprehensive benchmark designed for diverse quantity and layout control scenarios. Extensive experiments demonstrate that IMAGHarmony consistently outperforms existing state-of-the-art methods in both structural alignment and semantic accuracy. Notably, our framework achieves these superior results with remarkable efficiency, requiring only 200 training images and 10.6M trainable parameters, while maintaining robustness across different diffusion backbones.

\section*{Declarations}
\textbf{Funding} This work is supported by the Major Research Program of Jiangsu Province (Grant No. BG2024042). \\
\textbf{Competing Interests} The authors declare that they have no competing financial interests or personal relationships that could have appeared to influence the work reported in this paper. \\
\textbf{Data and Code Availability} The data, models, and code that support the findings of this study are available at \url{https://github.com/muzishen/IMAGHarmony}.

\bibliography{sn-bibliography}

\end{document}